\documentclass{article} 
\usepackage{iclr2027_conference,times}

\usepackage{amsmath,amsfonts,bm}

\def\eqref#1{equation~\ref{#1}}

\def\1{\bm{1}}

\DeclareMathAlphabet{\mathsfit}{\encodingdefault}{\sfdefault}{m}{sl}
\SetMathAlphabet{\mathsfit}{bold}{\encodingdefault}{\sfdefault}{bx}{n}

\usepackage{hyperref}
\usepackage{url}
\usepackage{graphicx}
\usepackage{xcolor}
\usepackage{booktabs}
\usepackage{multirow}
\graphicspath{{figures/}}
\usepackage{capt-of}
\usepackage{wrapfig}
\usepackage{enumitem}
\usepackage{placeins}
\usepackage{float}
\usepackage{listings}
\usepackage[most]{tcolorbox}
\usepackage[skip=4pt]{caption}  
\definecolor{promptbg}{HTML}{F6F7F9}
\definecolor{promptframe}{HTML}{4A5A6E}
\definecolor{promptslot}{HTML}{1F6FB2}
\definecolor{promptmark}{HTML}{B5471F}
\newcommand{\pslot}[1]{\textcolor{promptslot}{\textit{\textless#1\textgreater}}}
\newcommand{\pvar}[1]{\textcolor{promptslot}{\textit{#1}}}
\newcommand{\pmark}[1]{\textcolor{promptmark}{\textbf{#1}}}
\lstdefinestyle{prompt}{
  basicstyle=\ttfamily\footnotesize,
  breaklines=true,
  breakatwhitespace=true,
  columns=fullflexible,
  keepspaces=true,
  breakindent=0pt,
  aboveskip=0pt,
  belowskip=0pt,
  escapeinside={(*}{*)},
  literate={<}{\textless}1 {>}{\textgreater}1,
}
\newtcblisting{promptbox}[1]{
  enhanced, breakable, listing only,
  listing options={style=prompt},
  colback=promptbg, colframe=promptframe, colbacktitle=promptframe,
  coltitle=white, fonttitle=\small\sffamily\bfseries,
  boxrule=0.6pt, arc=2pt, left=6pt, right=6pt, top=4pt, bottom=4pt,
  title={#1},
}

\newcommand{\method}{SimEX}
\newcommand{\degree}{^{\circ}}
\newcommand{\toolboxfile}{\texttt{toolbox.py}}
\newcommand{\skillmd}{\texttt{skill.md}}

\newcommand{\pcnt}[2]{\ensuremath{#1/#2}}
\newcommand{\bpcnt}[2]{{\boldmath\ensuremath{#1/#2}}}

\title{\method{}: Simulation-Integrated Robotics \\AutoResearch}

\newif\ifpreprint
\preprinttrue

\ifpreprint
  \iclrfinalcopy
  \author{Jiaheng Hu$^{1,2,*}$, Roberto Mart\'in-Mart\'in$^{2}$, Peter Stone$^{2}$, Rocky Duan$^{1}$, \\
  \textbf{Zhenyu Jiang$^{1,\dagger}$, Guanya Shi$^{1,3,\dagger}$} \\
  $^1$Amazon FAR (Frontier AI \& Robotics) \quad
  $^2$The University of Texas at Austin \\
  $^3$Carnegie Mellon University}
\else
  \author{Anonymous authors \\
  Paper under double-blind review}
\fi
\begin{document}

\maketitle
\ifpreprint\lhead{}\renewcommand{\headrulewidth}{0pt}
  {\renewcommand{\thefootnote}{\fnsymbol{footnote}}%
   \footnotetext[1]{Work done during internship at Amazon FAR.}%
   \footnotetext[2]{Amazon FAR team co-lead.}}
\fi

\begin{abstract}

Coding agents powered by large language models (LLMs) have shown remarkable
abilities to autonomously reason about and achieve goals in the digital world.
However, bringing this success to the physical world remains challenging. On
the one hand, direct generation methods (e.g., Code as Policies) often suffer
from the LLMs' insufficient understanding of robots and physical environments.
On the other hand, iterative trial-and-error tuning in the physical world
(e.g., physical autoresearch) induces significant experimental cost and safety
concerns. We introduce \method{}: Simulation-Integrated Robotics AutoResearch, an autoresearch framework that tightly
integrates simulated experimentation, enabling coding agents to efficiently
acquire physical capabilities for controlling real robots. \method{} operates
in two stages.
First, the agent conducts open-ended \emph{probe-and-optimize} iterations in simulation, developing a robot toolbox with robust and generalizable capabilities. Second, the agent adapts the toolbox and the simulator together through only a
few physical trials: each trial corrects the simulator, and the corrected
simulator is used to diagnose failures and screen candidate repairs.
We evaluate \method{} extensively in sim-to-sim settings and on physical robots. On challenging real-world manipulation tasks including towel folding, barcode scanning, and plate manipulation, \method{} enables coding agents to efficiently acquire robot skills without any demonstration and with only 10 minutes of real-robot interaction. 
These results suggest that simulation can be a critical component in achieving physical intelligence, not only as a source of training data that must closely replicate the real world, but also as a roughly correct laboratory where a coding agent develops the knowledge and procedures needed to act on the robot. More details and robot videos at \url{https://robo-simex.github.io/}.
\end{abstract}

\section{Introduction}
\label{sec:intro}

Coding agents powered by large language models (LLMs) have emerged as capable
problem solvers in digital environments. Given access to a workspace, they can
write and debug software, resolve repository-level issues
\citep{jimenez2024swebench}, conduct scientific experiments
\citep{lu2024aiscientist}, and discover new mathematical constructions and
algorithms \citep{romera2024funsearch,novikov2025alphaevolve}. These
capabilities already enable coding agents to automate substantial amounts of
digital work. The next step toward realizing their full potential is to extend
them beyond computers and enable them to act through robots in the physical
world.

However, bringing coding agents into the physical world introduces challenges
that do not arise in purely digital environments. One popular approach
directly generates robot control code from a natural-language instruction (aka ``Code as Policies'')
\citep{liang2023code,singh2023progprompt}. 
While this approach produces a control program in a single pass, without any trial and
error on the robot, its performance often depends on carefully designed,
user-provided physical primitives, such as perception~\citep{kamath2021mdetr,gu2021open},
motion-planning~\citep{huang2023voxposer,zucker2013chomp}, and manipulation functions~\citep{sundermeyer2021contact}. 
Without these primitives, which are often embodiment- and environment-specific and costly to hand-engineer, the LLM must guess quantities that appear neither in its training data nor in the prompt, such as camera calibration errors or reachable grasps. These guesses rarely meet the tight tolerances of physical tasks, and often result in catastrophic failure of these direct generation methods.

Building on iterative autoresearch in software
\citep{karpathy2026autoresearch}, a second line of work lets coding agents
improve robot policies from physical feedback
\citep{xiao2026enpire}. Unlike direct generation, this process allows the
agent to learn from the consequences of its actions and correct mistakes that
could not be anticipated from the initial prompt. However, experimentation on
real robots is costly and potentially unsafe: each trial consumes robot time,
untested code can cause collisions, and noisy rollouts can provide misleading
signals for subsequent revisions. As a result, real-world autoresearch systems such as ENPIRE
\citep{xiao2026enpire} depend on carefully engineered physical harnesses,
including automatic scene resets, success verification, and safety monitoring,
to sustain long sequences of robot trials.

How can we bring coding agents onto robots without relying extensively on
physical trials? Our key idea is to use simulation as a bridge between the
coding agent and the physical world. Simulation is particularly well suited to
this role: on the one hand, it is defined and manipulated through code, making
it naturally accessible to a coding agent; on the other hand, it encodes
physical laws, robot dynamics, and workspace geometry needed to solve
real-world tasks. The agent can therefore repeat experiments and compare hypotheses safely in simulation before
acting on the robot. 
However, no simulator perfectly reproduces the physical world, and capabilities optimized purely in simulation can often exploit its inaccuracies. An effective approach must therefore tolerate an imperfect simulator and ground what it learns in a small number of real-world trials.

\begin{figure}[t]
\centering
\includegraphics[width=\linewidth]{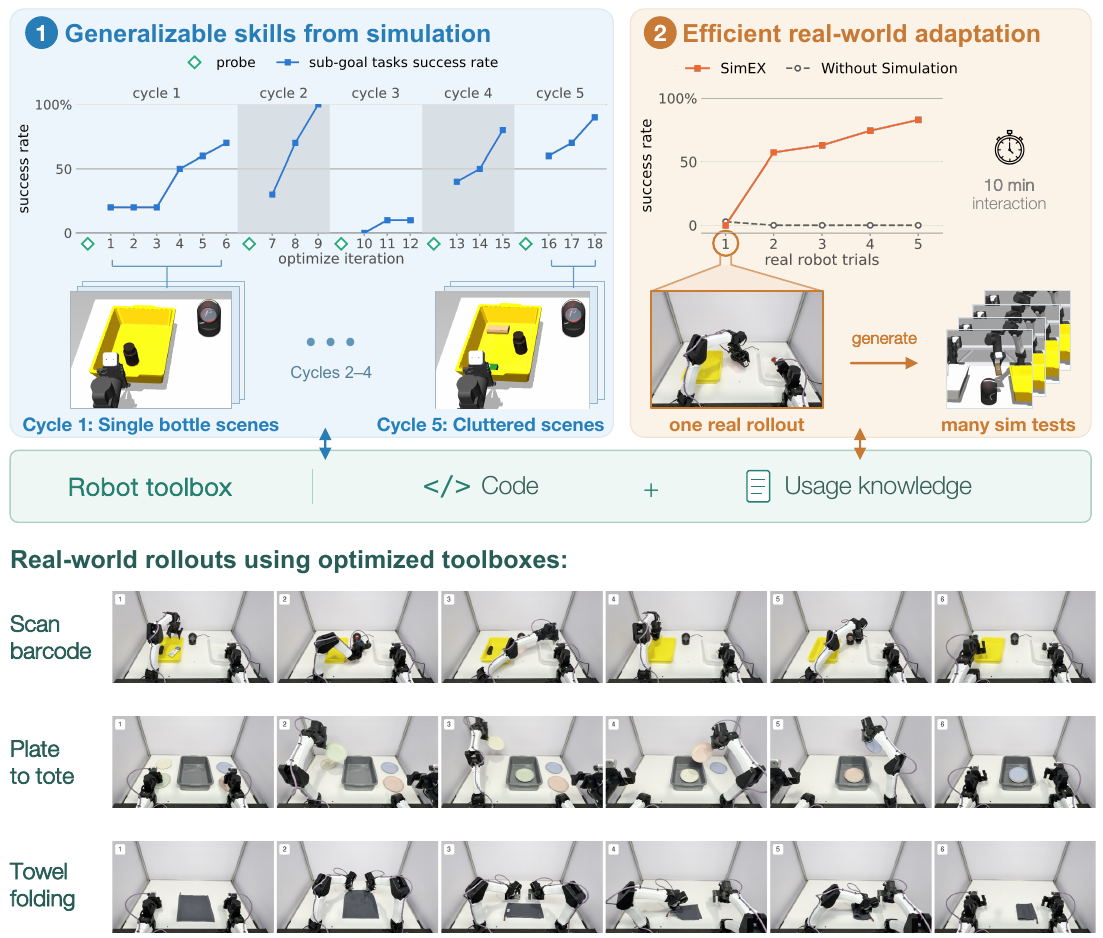}
\caption{\method{} brings coding agents to physical robots through simulation.
  (1)~The agent develops diverse robot capabilities through open-ended autoresearch in simulation. (2)~The agent then efficiently adapts to the real robot within a few trials, turning each real rollout into many simulated tests of candidate repairs. Bottom: physical-robot rollouts using the optimized toolboxes.}
\label{fig:teaser}
\end{figure}

We introduce \method{}, a simulation-integrated
autoresearch framework that enables a coding agent to efficiently acquire robot
capabilities.
Given an autonomously constructed simulation of the workspace, \method{}
follows a two-stage process (Figure~\ref{fig:overview}).
In Stage~1, the agent conducts open-ended autoresearch in simulation without a
fixed, human-specified evaluation metric. It probes the current toolbox on
self-generated tasks and workspace variations that expose its weaknesses,
then improves the toolbox through repeated simulated experiments. In Stage~2,
the agent adapts the resulting toolbox through a few
physical trials, using each robot rollout to generate improvement hypotheses
and returning to simulation to screen candidate repairs. Importantly, the two stages make \method{} robust to an imperfect simulator (e.g., the coarse deformable simulator for our towel folding experiments): open-ended exploration in Stage~1 prevents the toolbox from overfitting to any single simulated scenario, and real-world grounding in Stage~2 corrects what the simulator gets wrong. This distinguishes \method{} from real-to-sim-to-real methods~\citep{torne2024reconciling}, which use high-fidelity simulation to train one policy for one reconstructed scene; \method{} instead transfers knowledge and procedures, the limits it has measured and the strategies it has found to work, which need only a simulator that gets the qualitative physics right and therefore carry over across tasks.

We evaluate \method{} in sim-to-sim settings and on physical robots, across four types of coding agents (Fable 5.1, Opus 5, GPT-5.5 and GPT-6 Astra) and
three manipulation families spanning long-horizon object handling,
contact-rich manipulation, and deformable-object control. \method{} solves
these challenging tasks without demonstrations and substantially outperforms
approaches that directly generate robot code or adapt purely through physical
trials. On physical hardware, with only 10 minutes of on-robot interaction per
task, \method{} succeeds in 26 of 30 evaluation trials across the three task
families, while the strongest baseline succeeds in only 3.

To summarize, our contributions are:
\vspace{-0.4em} 
\begin{itemize}[leftmargin=2em,itemsep=0pt,parsep=0pt,topsep=2pt]
  \item We introduce a simulation-integrated autoresearch framework that enables coding agents to efficiently control physical robots.
  \item We develop a novel open-ended simulated autoresearch procedure
  (Stage~1),
  in which the agent generates tasks and environment variations without a
  ground-truth distribution or human-specified evaluation metric, and develops generalizable
  capabilities.
  \item We develop a sample-efficient physical adaptation procedure (Stage~2)
  that interleaves limited physical experience with simulated experimentation.
  \item We demonstrate that \method{} solves challenging long-horizon,
  contact-rich, and deformable-object manipulation tasks on physical robots
  without any demonstrations.
\end{itemize}

\section{Method}
\label{sec:method}

\begin{figure*}[t]
\centering
\includegraphics[width=\textwidth]{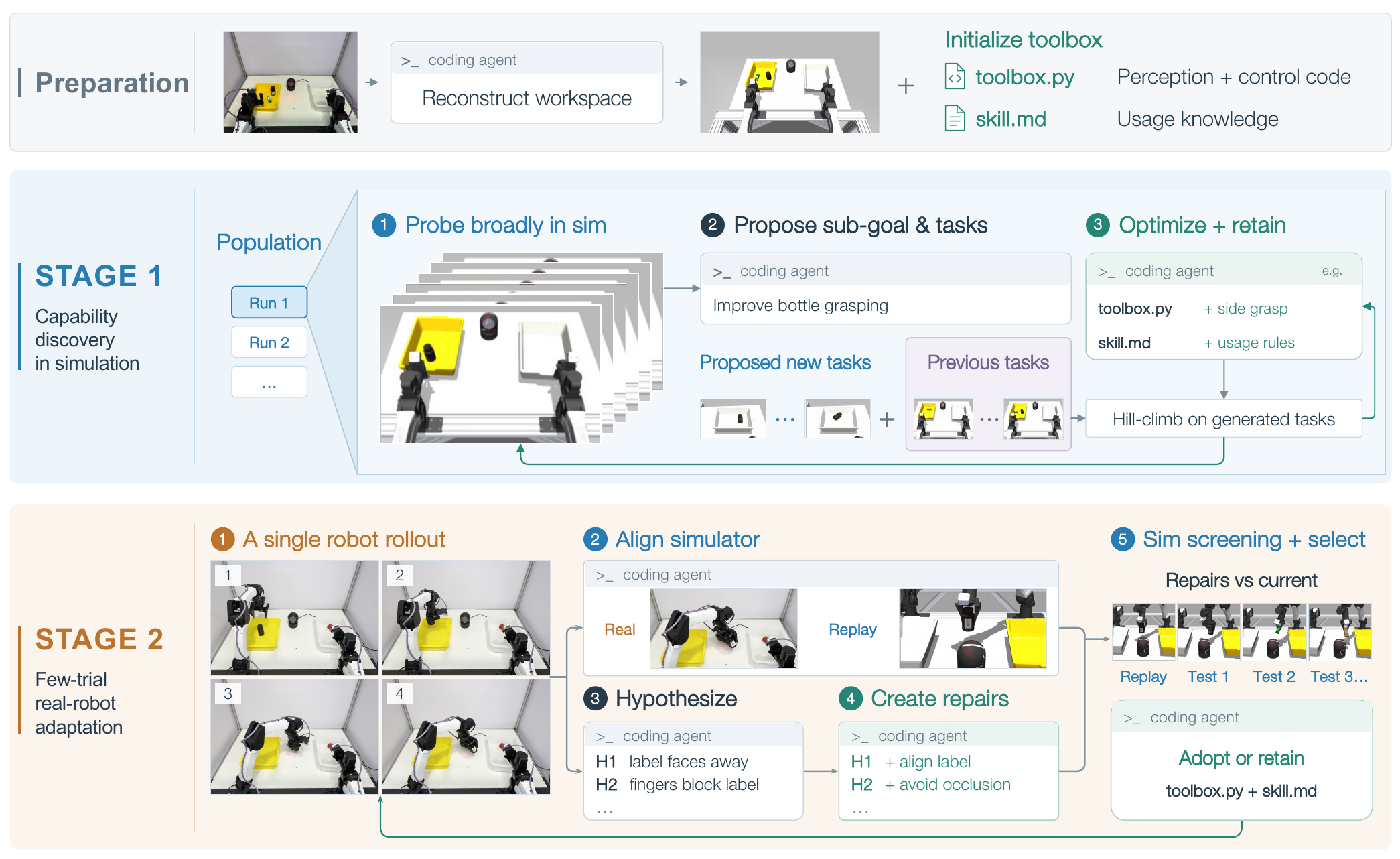}
\caption{Overview of \method{}. Preparation reconstructs the physical
workspace in simulation and initializes the toolbox. Stage~1 runs independent
probe--focus--optimize cycles, retaining improvements on focused tasks and
capability-bank replay. Stage~2 executes one robot rollout, aligns the
simulator through fixed-policy replay, creates candidate repairs, and returns simulated screening evidence to the coding
agent as supportive evidence for repair selection. }
\label{fig:overview}
\end{figure*}

\method{} aims to enable a coding agent to efficiently acquire reusable
capabilities for controlling physical robots. It begins by autonomously
constructing a simulation of the physical workspace
(Section~\ref{sec:prep}). \method{} then uses this simulation to improve a
portable robot toolbox through two stages. In Stage~1, \method{}
develops generalizable capabilities through open-ended simulated
experimentation, producing a toolbox that supports diverse tasks under
workspace variations (Section~\ref{sec:stage1}). In Stage~2,
\method{} adapts this toolbox with a small physical trial budget, returning to
simulation to screen repairs between trials
(Section~\ref{sec:stage2}). Importantly, this two-stage design
confines on-robot execution to a compact final phase after the agent has
already acquired an initial toolbox. This design has two benefits.
First, the initial capabilities make physical rollouts safer and more
efficient.
Second, concentrating these trials into one phase provides the practical
benefit of allowing a human operator to initialize and monitor the robot in
one session rather than repeatedly throughout the longer optimization process.
Figure~\ref{fig:overview} visualizes the full pipeline.

\subsection{Problem Setup}
\label{sec:library}

\method{} optimizes a portable robot capability library, which we call the
\emph{toolbox}. It comprises two files. \toolboxfile{}
implements perception and control primitives, while \skillmd{} documents how
to use them, including measured limits and known failure modes. When the agent
encounters a task, \method{} makes a fresh query to an LLM with the current
toolbox and a natural-language task description. Acting as a code-as-policy
model \citep{liang2023code}, the LLM writes an executable program that invokes
the primitives in \toolboxfile{} using the guidance in \skillmd{}
(prompt in Appendix~\ref{app:cap-prompt}). Together,
the LLM and toolbox can be viewed as a multi-task policy that maps a
natural-language task description to a visuomotor program.

\textbf{Toolbox boundary.}~The toolbox is an editable layer built above a fixed low-level robot API. For
scene perception, the toolbox has access only to raw RGBD images, and receives no privileged scene state. The robot API
also provides joint and end-effector proprioception,
inverse-kinematics and reachability queries, Cartesian end-effector or
absolute joint-position targets, and gripper commands, but no predefined
task-level manipulation skills. 
During optimization, \method{} improves this toolbox while the underlying robot API remains fixed.

\textbf{Agent-driven autoresearch.}~Both optimization stages run as autoresearch loops
\citep{karpathy2026autoresearch} driven entirely by a coding agent. Each stage
starts a coding-agent session with an \texttt{agent\_program.md} that states the
goal, the rules the agent must follow, the steps of one iteration, and the
harness commands available to it. The agent then drives the loop
autonomously. In each iteration, it chooses what to test and edits the toolbox
accordingly. It then launches simulated rollouts or robot trials through the
harness and records what the results teach it before the next iteration
begins. The harness only executes and scores rollouts, so every optimization
decision belongs to the agent. The loop ends when the stage's iteration budget
is spent.

Within this loop, the agent revises \toolboxfile{} and
\skillmd{} as a pair. The task-specific program generated via CaP is discarded after execution;
only the shared toolbox persists. Consequently, knowledge transfers across
iterations only through updates to the toolbox. The toolbox therefore stores
task-agnostic capabilities rather than a solution to one instruction, allowing
it to support many tasks and workspace variations.

\subsection{Preparation: Constructing the simulation sandbox}
\label{sec:prep}

\method{} starts by constructing a simulation sandbox from an image of the
physical workspace. The goal is to produce an executable approximation that
subsequent autoresearch can build on, rather than a precisely calibrated or
photorealistic digital twin. In principle, this initialization can be produced
by an off-the-shelf real-to-sim system, such as
SimFoundry or VIGA
\citep{ranawaka2026simfoundrymodularautomatedscene,yin2026visionasinversegraphicsagentinterleavedmultimodal}.
In our implementation, we simply give a frontier coding agent the high-level
objective of reconstructing the observed physical workspace in a target
simulation platform (e.g., mjlab). The agent has access to an empty YAM station
scene template and a predefined library of 3D meshes, and receives a workspace
image and a natural-language description of the objects present. It
autonomously selects and configures the assets. 

The reconstruction initializes three components: the simulator with the robot
and workspace geometry,
an initial \toolboxfile{} containing an initial perception module based on the input image,
and an initial \skillmd{} describing the fixed robot API. Together, these components provide an executable
starting point for the agent's optimization.

\subsection{Stage 1: Open-ended capability discovery in simulation}
\label{sec:stage1}

With this simulation initialization in place, the most direct approach would
be to optimize only for the requested task in the reconstructed simulator.
Doing so, however, risks overfitting to simulator-specific geometry and
dynamics that may not transfer to the physical system. Stage~1 instead asks
the coding agent to acquire broad capabilities before any robot trial by
repeatedly discovering and repairing weaknesses in the toolbox. The agent
generates both the tasks used to expose these weaknesses and the success
conditions used to evaluate them, rather than optimizing against a fixed task
distribution or human-specified metric.

Specifically, the coding agent first
probes the current toolbox with a broad set of self-generated tasks and
environment variations. It then selects one systematic weakness, constructs a
focused training set around it, and improves the toolbox on these tasks
together with replay tasks drawn from previously acquired capabilities. The
updated toolbox begins the next cycle, which probes broadly again to discover
a new weakness. \method{} runs several independent copies of this complete
cycle in parallel to diversify the resulting capabilities, where the copies can learn from each other. We describe these steps
in detail below. Figure~\ref{fig:overview} (blue panel) illustrates this
stage.

\textbf{Step 1: open-ended capability probing.}~The probing phase searches for capabilities that the current toolbox is
missing. The coding agent generates open-ended task and environment variations,
such as changes to the task goal, object categories, object physical
properties (e.g., mass and friction), object placements, or the robot's
initial configuration. For each configuration, it also writes an executable
success predicate over simulator state. These predicates make the
self-generated tasks automatically scorable without imposing one global
evaluation metric. The agent runs these configurations through the harness and
compares their successes and failures to identify a systematic limitation.
Probing therefore prioritizes discovering informative weaknesses over
maximizing performance on a predetermined task set. The outcome of probing is
a set of candidate weaknesses for the next step to choose from.

\textbf{Step 2: sub-goal selection.}~A broad probe set may reveal several unrelated failures. To produce a coherent
toolbox update, the coding agent selects one systematic weakness as the
sub-goal for the next optimization cycle and constructs a fixed set of
training tasks that isolates it. Fixing the sub-goal and
training tasks within one cycle provides a stable objective for comparing
successive toolbox revisions in the next step.

\textbf{Step 3: sub-goal-driven autoresearch.}~With the training set fixed, the agent proposes revisions to
\toolboxfile{}, \skillmd{}, or both and hill-climbs their performance on this
task set through repeated simulated experiments. The improved toolbox is then
probed again, which starts the next cycle. Repeating the full
probe--focus--optimize cycle over newly discovered sub-goals gradually expands
the toolbox's capability coverage. Two further mechanisms keep this loop from
narrowing: capability-bank replay guards against forgetting across cycles, and
population-based optimization diversifies the self-generated curriculum.

\textbf{Capability-bank replay.}~Focusing on one weakness creates the risk that an edit removes capabilities
learned in earlier cycles. To guard against such regressions, the coding agent
maintains a capability bank of previously solved tasks and samples replay
tasks from it during hill-climbing. Autoresearch therefore optimizes
the new sub-goal and previously acquired capabilities jointly. 

\textbf{Population-based optimization.}~Even repeated cycles can follow a narrow self-generated curriculum and/or get stuck in local optima. To
diversify the search, \method{} maintains a population of independent
optimization processes. Each member executes the full
probe--focus--optimize cycle with its own worktree, toolbox, generated task
sets, and capability bank, and terminates after a fixed exploration budget of
20 optimization iterations in our experiments. At probe boundaries, one
member may inspect a short description of another member's experiments. It may
reimplement an idea but cannot copy the other implementation. This exchange
allows useful hypotheses to spread while preserving diverse optimization
trajectories and independent implementations.

\subsection{Stage 2: Simulation-assisted few-trial adaptation on the robot}
\label{sec:stage2}

Although \method{} develops broad capabilities in Stage~1, directly porting
the resulting toolbox to a physical robot can still produce task failures
due to the sim-to-real gap. Stage~2 bridges this gap by using a small number of real-robot
rollouts to improve the toolbox.
The central idea of Stage~2 is to turn each scarce robot rollout into a batch
of simulated experiments. Rather than spending the next physical trial on the
first plausible edit, \method{} uses simulation to screen several candidate repairs, each
built from a distinct improvement hypothesis.

Each iteration begins with one rollout on the robot. The harness records the raw
execution and replays the executed policy in simulation. When necessary, the coding agent first updates the simulator until the replay reproduces the relevant deployment behavior. Next, the agent formulates distinct, testable hypotheses about how the toolbox can be improved, and implements several candidate repairs based on these hypotheses. Finally, the agent screens these repairs by running every candidate and the current toolbox through simulated
replays and test tasks. The resulting videos, traces, and scores are returned
to the coding agent as context. Based on this context, the
agent judges how much to trust each simulated test, and decides which repair, if any,
to use in the next robot rollout. Thus, one physical rollout supports many
offline experiments instead of one untested physical edit. We describe the
five steps of one iteration below. Figure~\ref{fig:overview} (orange panel)
illustrates this stage.

\textbf{Step 1: robot rollout.}~At the start of an iteration, the harness gives the current toolbox and task
description to the policy-writing model, which produces one script through the
same prompt used during Stage~1 (Appendix~\ref{app:cap-prompt}). The script is executed once. The harness
stores the camera videos, every image frame read by the policy, and the full
policy transcript, including the perception and motion information printed
during execution. These raw artifacts provide the evidence for subsequent
simulated experiments.\looseness=-1

\textbf{Step 2: replay-based simulator alignment.}~Immediately after the robot rollout is scored, the harness replays the exact
executed policy in simulation under the same task configuration. The coding
agent qualitatively compares the recorded real and simulated traces and uses
its own judgment to determine whether the simulator reproduces the behavior
relevant to the observed outcome. If it does not, the agent keeps updating the simulator and reruns the same fixed-policy replay. These corrections persist and are applied to subsequent screening rollouts. 

\textbf{Step 3 \& 4: hypothesis and candidate repair generation.}~In parallel, the agent diagnoses the rollout. A robot rollout reveals what happened but may support several
explanations for what should change. The coding agent therefore
formulates concrete hypotheses from the recorded evidence, each identifying a
toolbox limitation and predicting an observable improvement. 
Based on these hypotheses, the coding agent creates several copies of the current
toolbox and implements a distinct candidate repair in each one. The next step tests these predictions in simulation.\looseness=-1

\textbf{Step 5: repair screening.}~To test these predictions before the next robot trial, the agent launches a screen in which the harness runs each candidate and the current toolbox in two settings. It first replays the
executed robot script from the reconstructed trial state, holding the script
fixed. It then evaluates them on a small set of simulated test tasks designed
to expose their differences and check whether a local fix generalizes beyond
the observed failure. Importantly, screening does not rank the candidates or apply an
automatic selection rule. Instead, the coding agent receives the resulting
videos, traces, and scores as context, judges how much to trust each test given
any remaining mismatch between simulation and the robot, and selects a repair
or retains the current toolbox. The selected toolbox is used for the next
robot rollout, which begins the next iteration.

\section{Experiments}
\label{sec:experiments}

We evaluate \method{} on three families of challenging robot manipulation
tasks spanning long-horizon object handling, contact-rich manipulation, and
deformable-object control. The evaluation combines controlled sim-to-sim
comparisons with baselines, tests on a physical robot, ablations of each stage,
and comparisons across four coding agents.

Through these experiments, we seek to answer five questions:
(1)~How effectively does \method{} enable coding agents to solve challenging
manipulation tasks compared with alternative approaches
(Section~\ref{sec:exp-results})?
(2)~Does this advantage carry over to a physical robot
(Section~\ref{sec:exp-real})?
(3)~Which design choices within each stage account for its performance
(Section~\ref{sec:exp-ablations})?
(4)~Do these conclusions hold across different coding agents
(Section~\ref{sec:exp-coding-agents})?
(5)~How does synthesizing a program compare with using the coding agent itself
as an online policy (Section~\ref{sec:exp-deployment-efficiency})?

\subsection{Experimental setup}
\label{sec:exp-scenes}
\label{sec:exp-baselines}
\label{sec:exp-protocol}

\textbf{Tasks.}~We consider three manipulation task families that require different
capabilities (Figure~\ref{fig:scenes}). In \emph{plate to tote}, the robot
moves plates from a table into a tote; a plate lying flat cannot be picked up
with a straightforward top-down grasp, so the agent must discover a
geometry-aware strategy. In \emph{towel folding}, the robot brings a
deformable towel into a target state, such as shifted, compacted, or folded,
and the outcome is highly sensitive to grasp location and drag trajectory.
\emph{Barcode scanning} is a long-horizon task in which the robot presents the
label of each object to a fixed barcode reader before stowing it in a second
tote. All tasks use a dual-arm YAM robot with one overhead camera and two
wrist cameras. For each family, we define seven evaluation tasks that are held
out from the coding agent. Appendix~\ref{app:tasks} describes each family and
task in detail.\looseness=-1

\textbf{Evaluation protocol.}~For each task, every adaptive method receives on-robot five trials. After optimization, we freeze the resulting toolbox and evaluate it in five i.i.d.
rollouts, generating a fresh task program from the frozen toolbox for each
rollout. We repeat this process with five random seeds, yielding 25
evaluation trials per method--task pair. Non-adaptive baselines follow the same
five-seed protocol without making deployment-time updates. We report
successful trials over the 25 scheduled evaluation trials. Unless noted
otherwise, scene-level results are unweighted means of the seven task-level
success rates, and error bars show 95\% bootstrap confidence intervals that resample runs within each task. All coding-agent and policy-writing calls use Fable~5.1, with maximum reasoning effort in Stage~1
and medium reasoning effort in Stage~2, except in the cross-agent comparison
(Section~\ref{sec:exp-coding-agents}).
Table~\ref{tab:hyperparameters} summarizes the autoresearch and evaluation
budgets used throughout the experiments. We report the agent thinking time
and token usage of both stages in Appendix~\ref{app:compute-cost}.

\textbf{Baselines.}~We compare against Direct CaP~\citep{liang2023code},
ENPIRE$^*$~\citep{xiao2026enpire}, Zero-Shot Sim2Real, which follows
DrEureka~\citep{ma2024dreureka}, and ASPIRE$^*$~\citep{lu2026aspire}. We also
include two \method{} variants, \method{} (no discovery) and \method{} (direct
repair), which remove open-ended discovery in Stage~1 and sandbox-guided repair
in Stage~2, respectively. An asterisk marks a baseline that we reimplement
under our common interface and budget. Appendix~\ref{app:baselines} describes
each method and the controlled-comparison setup.

\begin{figure*}[!t]
\centering
\includegraphics[width=\textwidth]{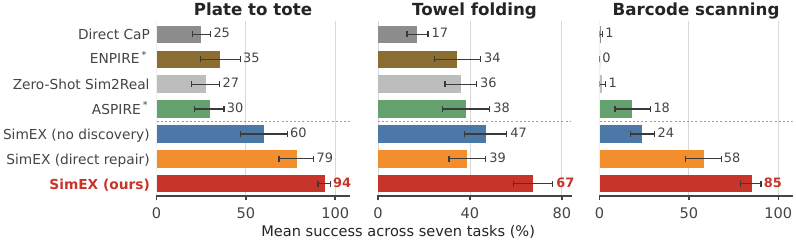}
\caption{Aggregate performance across the three evaluation task families.}
\label{fig:scene-baselines}
\end{figure*}

\subsection{Controlled sim-to-sim evaluation}
\label{sec:exp-results}

Comparing trial efficiency requires many repeated adaptation runs, which is
impractical to perform entirely on hardware. We therefore use an independently
implemented simulator as a stand-in for the physical robot. Barcode scanning
and plate to tote use MuJoCo \citep{todorov2012mujoco} as the
sandbox and Isaac Sim \citep{nvidia2025isaacsim} as the evaluation
environment. Towel folding uses FLASH \citep{luo2026flash} as the sandbox and
MJWarp Flex~\citep{mujoco_warp2025} as the evaluation environment. In each pair, the two simulators have
distinct physics implementations, contact solvers, assets, rendering
pipelines, and system parameters. The coding agent receives the same
observations and task outcomes available in the real robot setup, but cannot
inspect the evaluation simulator's state or source code. This setting
preserves a substantial transfer gap while supporting controlled repetition.\looseness=-1

We first summarize aggregate performance across the three task families in
Figure~\ref{fig:scene-baselines}, then report per-task success counts in
Table~\ref{tab:pertask-results} (Appendix~\ref{app:pertask}).
Figure~\ref{fig:scene-baselines} shows that \method{} achieves the highest
average success in all three task families. It also outperforms both stage-level
ablations in every family, supporting the contributions of open-ended
discovery in Stage~1 and sandbox-guided repair in Stage~2.
The per-task results show that this advantage is broad rather than driven by
a few tasks: \method{} matches or exceeds every external baseline on all 21
tasks. The separation is largest on barcode scanning, the longest-horizon
family, where \method{} reaches 85\% average success and no external baseline
exceeds 18\%.

\begin{figure*}[!t]
\centering
\begin{minipage}[t]{0.43\textwidth}
\centering
\includegraphics[width=\linewidth]{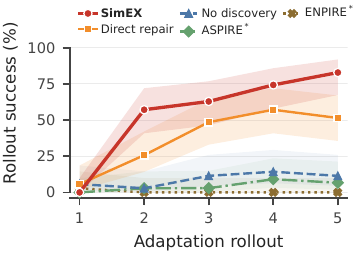}
\captionof{figure}{Online adaptation (Stage 2). Each
point is the average success rate of the deployment rollouts at that iteration.}
\label{fig:iteration-curve}
\end{minipage}
\hfill
\begin{minipage}[t]{0.54\textwidth}
\centering
\includegraphics[width=\linewidth]{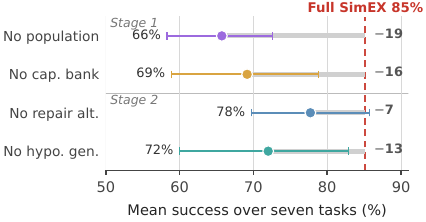}
\captionof{figure}{Fine-grained ablations. Points show mean task success, with the full \method{} result marked by the red
line. Every ablation reduces performance.}
\label{fig:ablations}
\end{minipage}
\end{figure*}

Figure~\ref{fig:iteration-curve} reports online adaptation (Stage 2) using the actual
deployment rollout from each iteration on the barcode task. The two methods that share the same
Stage~1 toolbox both begin near zero, so the toolbox does not transfer
directly. However, a single deployment rollout and one repair improve both
sharply, suggesting that Stage~1 still gives Stage~2 a strong starting point
for repair. Sandbox guidance more than doubles this gain, and the gap persists
through the fifth rollout, showing that the sandbox improves how effectively
limited deployment feedback is converted into repairs.

\subsection{Real-robot evaluation}
\label{sec:exp-real}

\begin{wraptable}{r}{0.39\textwidth}
\vspace{-12pt}
\captionsetup{justification=raggedright}
\caption{Physical-robot successful evaluation trials out of 10.}
\label{tab:real-robot-results}
\centering
\footnotesize
\setlength{\tabcolsep}{4pt}
\begin{tabular}{@{}lccc@{}}
\toprule
Method & Plate & Towel & Barcode \\
\midrule
Direct CaP         & \pcnt{0}{10} & \pcnt{0}{10} & \pcnt{0}{10} \\
ENPIRE$^*$         & \pcnt{2}{10} & \pcnt{1}{10} & \pcnt{0}{10} \\
ZS Sim2Real & \pcnt{0}{10} & \pcnt{2}{10} & \pcnt{0}{10} \\
ASPIRE$^*$         & \pcnt{2}{10} & \pcnt{0}{10} & \pcnt{0}{10} \\
\textbf{\method{} (ours)} & \bpcnt{10}{10} & \bpcnt{8}{10} & \bpcnt{8}{10} \\
\bottomrule
\end{tabular}
\vspace{-2pt}
\end{wraptable}
We next evaluate whether the same adaptation loop can operate on physical
hardware. We deploy each method on the physical YAM robot under common
task criteria; adaptive methods receive the same five-trial adaptation budget.
These five trials amount to only about 10 minutes of wall-clock robot
interaction per task.
Between each trial, a human operator resets the scene.
We select one representative task from each family: plate to tote, towel
folding, and barcode scanning. The selected tasks respectively require stowing
four plates, folding a towel in half, and scanning and stowing two objects.
We then evaluate each method in 10 physical trials per task.


As shown in Table~\ref{tab:real-robot-results}, \method{} succeeds in 26 of 30 physical
evaluation trials, while no baseline succeeds in more than 3. The advantage is
consistent across all three task families, including the two-object barcode
scanning task, on which every baseline fails all trials. These results show
that the capabilities developed in simulation and repaired through limited
physical feedback transfer across contact-rich, deformable, and long-horizon
real-world manipulation tasks.\looseness=-1 

Qualitatively, the adapted barcode-scanning policy
discovers an efficient single-arm strategy that grasps, presents, and stows
both objects without unnecessary bimanual coordination; the plate policy learned to change grasp side based on the distance between the plate and the tote; while the towel policy learned to flatten the towel after folding. 
Videos showcasing these real-robot trajectories
can be found on our website.\looseness=-1


\subsection{Ablations and further analysis}
\label{sec:exp-analysis}

\textbf{Ablation study.}~\label{sec:exp-ablations}%
The main comparison in Section~\ref{sec:exp-results} already ablates each
stage as a whole. Here, we ablate the mechanisms within each stage: removing the
population or capability-bank replay in Stage~1, and removing repair
alternatives or hypothesis generation in Stage~2
(Figure~\ref{fig:ablations}). Appendix~\ref{app:ablations} defines each
ablation and reports per-task counts. Results show that every ablation reduces performance. The largest drops arise
from the two Stage~1 ablations, removing population diversity and
capability-bank replay.

\textbf{Robustness across coding agents.}~\label{sec:exp-coding-agents}%
The main experiments use Fable~5.1 for both the optimization loop and
all policy-writing calls. To test whether our conclusions depend on this
agent, we repeat the complete seven-task barcode scanning comparison with three
alternatives: Opus~5, GPT-5.5, and GPT-6 Astra each replaces Fable~5.1 as both
the optimization agent and policy-writing model. All agents use their highest
reasoning effort in Stage~1 and medium reasoning effort in Stage~2.

\begin{wrapfigure}{r}{0.47\textwidth}
\vspace{-8pt}
\centering
\includegraphics[width=\linewidth]{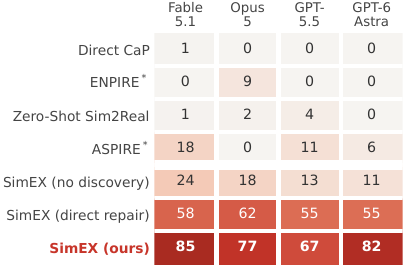}
\caption{Cross-agent comparison. Our main conclusion holds with different coding agents.}
\label{fig:cross-agent-heatmap}
\vspace{-8pt}
\end{wrapfigure}
Our conclusion is stable across coding agents. With every agent, \method{} is
the strongest method and leads the best external baseline by more than 50
percentage points
(Figure~\ref{fig:cross-agent-heatmap}; Appendix~\ref{app:cross-agent-results}, Figure~\ref{fig:cross-agent-gap}). Interestingly, absolute performance does depend on the strength of the coding agent. Under otherwise
identical end-to-end settings, replacing GPT-5.5 with GPT-6 Astra raises
\method{} from 67\% to 82\%. This result suggests that improvements in
coding-agent reasoning can translate into stronger robot performance, allowing our framework to benefit from future model progress. Per-task results for
all four configurations appear in
Appendix~\ref{app:cross-agent-results}.\looseness=-1

\textbf{Comparison with online agent control.}~\label{sec:exp-deployment-efficiency}%
An alternative to synthesizing a robot program is to use the coding agent
itself as an online policy~\citep{zhang2026unexpectedrobotpolicyearly}: after
each observation, the agent emits the next low-level action. We compare the two
deployment modes on three barcode scanning tasks
(Appendix~\ref{app:online-control}). \method{} makes a single model call of
about 30 seconds per task before execution, and all three generated programs
complete their validation rollout. The online policy instead waits over 50
times longer for model responses than it spends executing actions, and none of
its rollouts completes the task within the allotted wall-clock budget. Moving
coding-agent inference out of the execution loop therefore makes \method{}
substantially more practical for robotics.

\FloatBarrier

\section{Related Work}
\label{sec:related}

\method{} is most closely related to four lines of work: LLM-based robot
control~\citep{liang2023code,huang2022inner,vemprala2024chatgpt}, agentic
optimization~\citep{karpathy2026autoresearch,ma2024eureka,lu2026aspire,xiao2026enpire},
generative simulation and environment
design~\citep{dennis2020emergent,wang2023robogen,wang2024gensim}, and
sim-to-real transfer~\citep{tobin2017domain,ma2024dreureka,kumar2021rma}.
Unlike these works, \method{} uses simulation as a laboratory in which a coding
agent builds a reusable code toolbox, and returns to simulation between
physical trials to diagnose failures and screen repairs. We discuss each line
of work in detail in Appendix~\ref{app:related}.

\section{Conclusion}
\label{sec:conclusion}

We introduced \method{}, a framework that uses simulation as a bridge between
coding agents and physical robots. \method{} acquires broad capabilities
through open-ended simulated autoresearch in Stage~1, then uses limited
physical experience in Stage~2 to identify and repair transfer failures.
Across controlled sim-to-sim benchmarks and physical-robot tasks, this design
outperforms direct generation, deployment-only iteration,
simulation-only transfer, and simpler simulated search, while remaining robust
across four coding agents. More broadly, our results show that simulation is useful not only for direct policy transfer, but also as a laboratory in which a coding
agent iteratively develops and repairs robot capabilities before
returning to the physical world. This transfer happens at the level of knowledge and procedures rather than a policy, which is what allows a roughly correct simulator to be enough. One limitation is that our evaluation uses a
single robot platform; an important direction for future work is to evaluate
\method{} on additional robot platforms. In addition, a human operator still
resets the scene between physical trials in Stage~2; future work that automates
these resets would make \method{} fully autonomous.\looseness=-1

\bibliography{iclr2027_conference}

\begin{thebibliography}{50}
\providecommand{\natexlab}[1]{#1}
\providecommand{\url}[1]{\texttt{#1}}
\expandafter\ifx\csname urlstyle\endcsname\relax
  \providecommand{\doi}[1]{doi: #1}\else
  \providecommand{\doi}{doi: \begingroup \urlstyle{rm}\Url}\fi

\bibitem[Chebotar et~al.(2019)Chebotar, Handa, Makoviychuk, Macklin, Issac, Ratliff, and Fox]{chebotar2019closing}
Yevgen Chebotar, Ankur Handa, Viktor Makoviychuk, Miles Macklin, Jan Issac, Nathan Ratliff, and Dieter Fox.
\newblock Closing the sim-to-real loop: Adapting simulation randomization with real world experience.
\newblock In \emph{2019 international conference on robotics and automation (ICRA)}, pp.\  8973--8979. IEEE, 2019.

\bibitem[Chen et~al.(2026)Chen, Bai, Cao, Zeng, Lin, Lin, Liang, Ma, Huang, and Shou]{chen2026show}
Yanzhe Chen, Zechen Bai, Zhijun Cao, Wenzheng Zeng, Kevin~Qinghong Lin, Yiqi Lin, Guoqiang Liang, Kevin~Yuchen Ma, Qiming Huang, and Mike~Zheng Shou.
\newblock Show-harness: Just a vlm agent can play robots.
\newblock \emph{arXiv preprint arXiv:2609.10522}, 2026.

\bibitem[Dai et~al.(2024)Dai, Wong, Jiang, Wang, Gokmen, Zhang, Wu, and Fei-Fei]{dai2024automated}
Tianyuan Dai, Josiah Wong, Yunfan Jiang, Chen Wang, Cem Gokmen, Ruohan Zhang, Jiajun Wu, and Li~Fei-Fei.
\newblock Automated creation of digital cousins for robust policy learning.
\newblock \emph{arXiv preprint arXiv:2410.07408}, 2024.

\bibitem[Dennis et~al.(2020)Dennis, Jaques, Vinitsky, Bayen, Russell, Critch, and Levine]{dennis2020emergent}
Michael Dennis, Natasha Jaques, Eugene Vinitsky, Alexandre Bayen, Stuart Russell, Andrew Critch, and Sergey Levine.
\newblock Emergent complexity and zero-shot transfer via unsupervised environment design.
\newblock \emph{Advances in neural information processing systems}, 33:\penalty0 13049--13061, 2020.

\bibitem[Fu et~al.(2026)Fu, Yu, El-Refai, Kou, Xue, Huang, Xiao, Wang, Niu, Li, et~al.]{fu2026cap}
Letian Fu, Justin Yu, Karim El-Refai, Ethan Kou, Haoru Xue, Huang Huang, Wenli Xiao, Guanzhi Wang, Dantong Niu, Fei-Fei Li, et~al.
\newblock Cap-x: A framework for benchmarking and improving coding agents for robot manipulation.
\newblock \emph{arXiv preprint arXiv:2603.22435}, 2026.

\bibitem[{Google DeepMind} \& {NVIDIA}(2025){Google DeepMind} and {NVIDIA}]{mujoco_warp2025}
{Google DeepMind} and {NVIDIA}.
\newblock Mujoco warp (mjwarp): A gpu-optimized implementation of mujoco in nvidia warp, 2025.
\newblock URL \url{https://github.com/google-deepmind/mujoco_warp}.

\bibitem[Gu et~al.(2021)Gu, Lin, Kuo, and Cui]{gu2021open}
Xiuye Gu, Tsung-Yi Lin, Weicheng Kuo, and Yin Cui.
\newblock Open-vocabulary object detection via vision and language knowledge distillation.
\newblock \emph{arXiv preprint arXiv:2104.13921}, 2021.

\bibitem[Ha et~al.(2023)Ha, Florence, and Song]{ha2023scaling}
Huy Ha, Pete Florence, and Shuran Song.
\newblock Scaling up and distilling down: Language-guided robot skill acquisition.
\newblock In \emph{Conference on robot learning}, pp.\  3766--3777. PMLR, 2023.

\bibitem[Hu et~al.(2025)Hu, Stone, and Martín-Martín]{hu2025slac}
Jiaheng Hu, Peter Stone, and Roberto Martín-Martín.
\newblock {SLAC}: Safe and efficient real-robot reinforcement learning via unsupervised simulation pre-training, 2025.
\newblock URL \url{https://arxiv.org/abs/2506.04147}.

\bibitem[Hu et~al.(2026)Hu, Shridhar, Lu, Shah, Chiang, Tan, and Xie]{hu2026mattersorchestratingrobotpolicies}
Jiaheng Hu, Mohit Shridhar, Caden Lu, Dhruv Shah, Hao-Tien~Lewis Chiang, Jie Tan, and Annie Xie.
\newblock What matters in orchestrating robot policies: A systematic study of hierarchical {VLA} agents, 2026.
\newblock URL \url{https://arxiv.org/abs/2606.10267}.

\bibitem[Huang et~al.(2023{\natexlab{a}})Huang, Wang, Zhang, Li, Wu, and Fei-Fei]{huang2023voxposer}
Wenlong Huang, Chen Wang, Ruohan Zhang, Yunzhu Li, Jiajun Wu, and Li~Fei-Fei.
\newblock Voxposer: Composable 3d value maps for robotic manipulation with language models.
\newblock \emph{arXiv preprint arXiv:2307.05973}, 2023{\natexlab{a}}.

\bibitem[Huang et~al.(2023{\natexlab{b}})Huang, Xia, Xiao, Chan, Liang, Florence, Zeng, Tompson, Mordatch, Chebotar, Sermanet, Jackson, Brown, Luu, Levine, Hausman, and Ichter]{huang2022inner}
Wenlong Huang, Fei Xia, Ted Xiao, Harris Chan, Jacky Liang, Pete Florence, Andy Zeng, Jonathan Tompson, Igor Mordatch, Yevgen Chebotar, Pierre Sermanet, Tomas Jackson, Noah Brown, Linda Luu, Sergey Levine, Karol Hausman, and Brian Ichter.
\newblock Inner monologue: Embodied reasoning through planning with language models.
\newblock In Karen Liu, Dana Kulic, and Jeff Ichnowski (eds.), \emph{Proceedings of The 6th Conference on Robot Learning}, volume 205 of \emph{Proceedings of Machine Learning Research}, pp.\  1769--1782. PMLR, 14--18 Dec 2023{\natexlab{b}}.
\newblock URL \url{https://proceedings.mlr.press/v205/huang23c.html}.

\bibitem[Jia et~al.(2026)Jia, Lin, Zhang, Zhang, Liu, and Jiang]{jia2026agent}
Mengzhao Jia, Yang Lin, Xixin Zhang, Zhihan Zhang, Xiaobai Liu, and Meng Jiang.
\newblock Agent as policy for robotic manipulation.
\newblock \emph{arXiv preprint arXiv:2609.12541}, 2026.

\bibitem[Jiang et~al.(2021)Jiang, Grefenstette, and Rockt{\"a}schel]{jiang2021prioritized}
Minqi Jiang, Edward Grefenstette, and Tim Rockt{\"a}schel.
\newblock Prioritized level replay.
\newblock In \emph{International Conference on Machine Learning}, pp.\  4940--4950. PMLR, 2021.

\bibitem[Jimenez et~al.(2024)Jimenez, Yang, Wettig, Yao, Pei, Press, and Narasimhan]{jimenez2024swebench}
Carlos~E Jimenez, John Yang, Alexander Wettig, Shunyu Yao, Kexin Pei, Ofir Press, and Karthik~R Narasimhan.
\newblock {SWE}-bench: Can language models resolve real-world {Github} issues?
\newblock In \emph{The Twelfth International Conference on Learning Representations}, 2024.
\newblock URL \url{https://openreview.net/forum?id=VTF8yNQM66}.

\bibitem[Kamath et~al.(2021)Kamath, Singh, LeCun, Synnaeve, Misra, and Carion]{kamath2021mdetr}
Aishwarya Kamath, Mannat Singh, Yann LeCun, Gabriel Synnaeve, Ishan Misra, and Nicolas Carion.
\newblock Mdetr-modulated detection for end-to-end multi-modal understanding.
\newblock In \emph{2021 IEEE/CVF international conference on computer vision (ICCV)}, pp.\  1760--1770. IEEE, 2021.

\bibitem[Karpathy(2026)]{karpathy2026autoresearch}
Andrej Karpathy.
\newblock autoresearch.
\newblock GitHub repository, 2026.
\newblock URL \url{https://github.com/karpathy/autoresearch}.

\bibitem[Kumar et~al.(2021)Kumar, Fu, Pathak, and Malik]{kumar2021rma}
Ashish Kumar, Zipeng Fu, Deepak Pathak, and Jitendra Malik.
\newblock Rma: Rapid motor adaptation for legged robots.
\newblock \emph{arXiv preprint arXiv:2107.04034}, 2021.

\bibitem[Liang et~al.(2023)Liang, Huang, Xia, Xu, Hausman, Ichter, Florence, and Zeng]{liang2023code}
Jacky Liang, Wenlong Huang, Fei Xia, Peng Xu, Karol Hausman, Brian Ichter, Pete Florence, and Andy Zeng.
\newblock Code as policies: Language model programs for embodied control.
\newblock In \emph{2023 IEEE International Conference on Robotics and Automation (ICRA)}, pp.\  9493--9500. IEEE, May 2023.
\newblock \doi{10.1109/ICRA48891.2023.10160591}.
\newblock URL \url{https://doi.org/10.1109/ICRA48891.2023.10160591}.

\bibitem[Liu et~al.(2024)Liu, Fang, Abbeel, and Levine]{liu2024moka}
Fangchen Liu, Kuan Fang, Pieter Abbeel, and Sergey Levine.
\newblock Moka: Open-vocabulary robotic manipulation through mark-based visual prompting.
\newblock In \emph{First Workshop on Vision-Language Models for Navigation and Manipulation at ICRA 2024}, 2024.

\bibitem[Lu et~al.(2024)Lu, Lu, Lange, Foerster, Clune, and Ha]{lu2024aiscientist}
Chris Lu, Cong Lu, Robert~Tjarko Lange, Jakob Foerster, Jeff Clune, and David Ha.
\newblock The {AI} scientist: Towards fully automated open-ended scientific discovery, 2024.
\newblock URL \url{https://arxiv.org/abs/2408.06292}.

\bibitem[Lu et~al.(2026)Lu, Wu, Kou, Fu, Xiao, Mandlekar, Xu, Shi, Goldberg, Chen, Chowdhury, Zhu, Fan, and Wang]{lu2026aspire}
Runyu Lu, Yubo Wu, Ethan Kou, Letian Fu, Wenli Xiao, Ajay Mandlekar, Yinzhen Xu, Guanya Shi, Ken Goldberg, Ang Chen, Mosharaf Chowdhury, Yuke Zhu, Linxi~"Jim" Fan, and Guanzhi Wang.
\newblock {ASPIRE}: Agentic /skills discovery for robotics, 2026.
\newblock URL \url{https://arxiv.org/abs/2607.00272}.

\bibitem[Luo et~al.(2026)Luo, Zhou, Zhang, Liu, Huang, Yang, Han, Hu, Yang, Yu, Zeng, and Shi]{luo2026flash}
Siyuan Luo, Bingyang Zhou, Chong Zhang, Xin Liu, Zhenhao Huang, Gang Yang, Zhengtao Han, Xiaotian Hu, Eric Yang, Rymon Yu, Ziqiu Zeng, and Fan Shi.
\newblock {FLASH}: Fast learning via {GPU}-accelerated simulation for high-fidelity deformable manipulation in minutes, 2026.
\newblock URL \url{https://arxiv.org/abs/2604.17513}.

\bibitem[Ma et~al.(2024{\natexlab{a}})Ma, Liang, Wang, Huang, Bastani, Jayaraman, Zhu, Fan, and Anandkumar]{ma2024eureka}
Yecheng~Jason Ma, William Liang, Guanzhi Wang, De-An Huang, Osbert Bastani, Dinesh Jayaraman, Yuke Zhu, Linxi Fan, and Anima Anandkumar.
\newblock {Eureka}: Human-level reward design via coding large language models.
\newblock In \emph{The Twelfth International Conference on Learning Representations}, 2024{\natexlab{a}}.
\newblock URL \url{https://openreview.net/forum?id=IEduRUO55F}.

\bibitem[Ma et~al.(2024{\natexlab{b}})Ma, Liang, Wang, Wang, Zhu, Fan, Bastani, and Jayaraman]{ma2024dreureka}
Yecheng~Jason Ma, William Liang, Hung-Ju Wang, Sam Wang, Yuke Zhu, Linxi Fan, Osbert Bastani, and Dinesh Jayaraman.
\newblock Dreureka: Language model guided sim-to-real transfer, 2024{\natexlab{b}}.
\newblock URL \url{https://arxiv.org/abs/2406.01967}.

\bibitem[Mu et~al.(2024)Mu, Chen, Zhang, Chen, Yu, Ge, Chen, Liang, Hu, Tao, et~al.]{mu2024robocodex}
Yao Mu, Junting Chen, Qinglong Zhang, Shoufa Chen, Qiaojun Yu, Chongjian Ge, Runjian Chen, Zhixuan Liang, Mengkang Hu, Chaofan Tao, et~al.
\newblock Robocodex: Multimodal code generation for robotic behavior synthesis.
\newblock \emph{arXiv preprint arXiv:2402.16117}, 2024.

\bibitem[Novikov et~al.(2025)Novikov, Vũ, Eisenberger, Dupont, Huang, Wagner, Shirobokov, Kozlovskii, Ruiz, Mehrabian, Kumar, See, Chaudhuri, Holland, Davies, Nowozin, Kohli, and Balog]{novikov2025alphaevolve}
Alexander Novikov, Ngân Vũ, Marvin Eisenberger, Emilien Dupont, Po-Sen Huang, Adam~Zsolt Wagner, Sergey Shirobokov, Borislav Kozlovskii, Francisco J.~R. Ruiz, Abbas Mehrabian, M.~Pawan Kumar, Abigail See, Swarat Chaudhuri, George Holland, Alex Davies, Sebastian Nowozin, Pushmeet Kohli, and Matej Balog.
\newblock {AlphaEvolve}: A coding agent for scientific and algorithmic discovery, 2025.
\newblock URL \url{https://arxiv.org/abs/2506.13131}.

\bibitem[{NVIDIA}(2025)]{nvidia2025isaacsim}
{NVIDIA}.
\newblock {NVIDIA Isaac Sim}, 2025.
\newblock URL \url{https://docs.isaacsim.omniverse.nvidia.com/5.1.0/index.html}.
\newblock Version 5.1.

\bibitem[Parker-Holder et~al.(2022)Parker-Holder, Jiang, Dennis, Samvelyan, Foerster, Grefenstette, and Rockt{\"a}schel]{parker2022evolving}
Jack Parker-Holder, Minqi Jiang, Michael Dennis, Mikayel Samvelyan, Jakob Foerster, Edward Grefenstette, and Tim Rockt{\"a}schel.
\newblock Evolving curricula with regret-based environment design.
\newblock In \emph{International Conference on Machine Learning}, pp.\  17473--17498. PMLR, 2022.

\bibitem[Ranawaka et~al.(2026)Ranawaka, Wong, Pai, Chu, Dai, Moghani, Yin, Jiang, Durbano, Huynh, Fang, Xu, Zhang, Fei-Fei, Fan, Wen, Mandlekar, and Zhu]{ranawaka2026simfoundrymodularautomatedscene}
Nadun Ranawaka, Josiah Wong, Wei-Lin Pai, Wei-Teng Chu, Tianyuan Dai, Masoud Moghani, Hang Yin, Yunfan Jiang, Wesley Durbano, Brandon Huynh, Yu~Fang, Danfei Xu, Ruohan Zhang, Li~Fei-Fei, Linxi Fan, Bowen Wen, Ajay Mandlekar, and Yuke Zhu.
\newblock Simfoundry: Modular and automated scene generation for policy learning and evaluation, 2026.
\newblock URL \url{https://arxiv.org/abs/2606.28276}.

\bibitem[Romera-Paredes et~al.(2024)Romera-Paredes, Barekatain, Novikov, Balog, Kumar, Dupont, Ruiz, Ellenberg, Wang, Fawzi, Kohli, and Fawzi]{romera2024funsearch}
Bernardino Romera-Paredes, Mohammadamin Barekatain, Alexander Novikov, Matej Balog, M.~Pawan Kumar, Emilien Dupont, Francisco J.~R. Ruiz, Jordan~S. Ellenberg, Pengming Wang, Omar Fawzi, Pushmeet Kohli, and Alhussein Fawzi.
\newblock Mathematical discoveries from program search with large language models.
\newblock \emph{Nature}, 625\penalty0 (7995):\penalty0 468--475, Jan 2024.
\newblock ISSN 1476-4687.
\newblock \doi{10.1038/s41586-023-06924-6}.
\newblock URL \url{https://doi.org/10.1038/s41586-023-06924-6}.

\bibitem[Singh et~al.(2023)Singh, Blukis, Mousavian, Goyal, Xu, Tremblay, Fox, Thomason, and Garg]{singh2023progprompt}
Ishika Singh, Valts Blukis, Arsalan Mousavian, Ankit Goyal, Danfei Xu, Jonathan Tremblay, Dieter Fox, Jesse Thomason, and Animesh Garg.
\newblock {ProgPrompt}: Generating situated robot task plans using large language models.
\newblock In \emph{2023 IEEE International Conference on Robotics and Automation (ICRA)}, pp.\  11523--11530. IEEE, May 2023.
\newblock \doi{10.1109/ICRA48891.2023.10161317}.
\newblock URL \url{https://doi.org/10.1109/ICRA48891.2023.10161317}.

\bibitem[Sundermeyer et~al.(2021)Sundermeyer, Mousavian, Triebel, and Fox]{sundermeyer2021contact}
Martin Sundermeyer, Arsalan Mousavian, Rudolph Triebel, and Dieter Fox.
\newblock Contact-graspnet: Efficient 6-dof grasp generation in cluttered scenes.
\newblock In \emph{2021 IEEE international conference on robotics and automation (ICRA)}, pp.\  13438--13444. IEEE, 2021.

\bibitem[Tobin et~al.(2017)Tobin, Fong, Ray, Schneider, Zaremba, and Abbeel]{tobin2017domain}
Josh Tobin, Rachel Fong, Alex Ray, Jonas Schneider, Wojciech Zaremba, and Pieter Abbeel.
\newblock Domain randomization for transferring deep neural networks from simulation to the real world.
\newblock In \emph{2017 IEEE/RSJ International Conference on Intelligent Robots and Systems (IROS)}, pp.\  23--30. IEEE, Sept 2017.
\newblock \doi{10.1109/IROS.2017.8202133}.
\newblock URL \url{https://doi.org/10.1109/IROS.2017.8202133}.

\bibitem[Todorov et~al.(2012)Todorov, Erez, and Tassa]{todorov2012mujoco}
Emanuel Todorov, Tom Erez, and Yuval Tassa.
\newblock {MuJoCo}: A physics engine for model-based control.
\newblock In \emph{2012 IEEE/RSJ International Conference on Intelligent Robots and Systems}, pp.\  5026--5033. IEEE, Oct 2012.
\newblock \doi{10.1109/IROS.2012.6386109}.
\newblock URL \url{https://doi.org/10.1109/IROS.2012.6386109}.

\bibitem[Torne et~al.(2024)Torne, Simeonov, Li, Chan, Chen, Gupta, and Agrawal]{torne2024reconciling}
Marcel Torne, Anthony Simeonov, Zechu Li, April Chan, Tao Chen, Abhishek Gupta, and Pulkit Agrawal.
\newblock Reconciling reality through simulation: A real-to-sim-to-real approach for robust manipulation.
\newblock \emph{arXiv preprint arXiv:2403.03949}, 2024.

\bibitem[Vemprala et~al.(2024)Vemprala, Bonatti, Bucker, and Kapoor]{vemprala2024chatgpt}
Sai~H Vemprala, Rogerio Bonatti, Arthur Bucker, and Ashish Kapoor.
\newblock Chatgpt for robotics: Design principles and model abilities.
\newblock \emph{Ieee Access}, 12:\penalty0 55682--55696, 2024.

\bibitem[Wang et~al.(2023{\natexlab{a}})Wang, Xie, Jiang, Mandlekar, Xiao, Zhu, Fan, and Anandkumar]{wang2023voyager}
Guanzhi Wang, Yuqi Xie, Yunfan Jiang, Ajay Mandlekar, Chaowei Xiao, Yuke Zhu, Linxi Fan, and Anima Anandkumar.
\newblock {Voyager}: An open-ended embodied agent with large language models, 2023{\natexlab{a}}.
\newblock URL \url{https://arxiv.org/abs/2305.16291}.

\bibitem[Wang et~al.(2024)Wang, Ling, Yuan, Shridhar, Bao, Qin, Wang, Xu, and Wang]{wang2024gensim}
Lirui Wang, Yiyang Ling, Zhecheng Yuan, Mohit Shridhar, Chen Bao, Yuzhe Qin, Bailin Wang, Huazhe Xu, and Xiaolong Wang.
\newblock Gensim: Generating robotic simulation tasks via large language models.
\newblock In \emph{International Conference on Learning Representations}, volume 2024, pp.\  4890--4924, 2024.

\bibitem[Wang et~al.(2023{\natexlab{b}})Wang, Xian, Chen, Wang, Wang, Fragkiadaki, Erickson, Held, and Gan]{wang2023robogen}
Yufei Wang, Zhou Xian, Feng Chen, Tsun-Hsuan Wang, Yian Wang, Katerina Fragkiadaki, Zackory Erickson, David Held, and Chuang Gan.
\newblock Robogen: Towards unleashing infinite data for automated robot learning via generative simulation.
\newblock \emph{arXiv preprint arXiv:2311.01455}, 2023{\natexlab{b}}.

\bibitem[Xiao et~al.(2026)Xiao, Xie, Zhang, Lin, Fu, Xue, Lu, Yang, Dai, Wang, Wu, Wang, Sastry, Goldberg, Fan, Zhu, and Shi]{xiao2026enpire}
Wenli Xiao, Jia Xie, Tonghe Zhang, Haotian Lin, Letian~"Max" Fu, Haoru Xue, Jalen Lu, Yi~Yang, Cunxi Dai, Zi~Wang, Jimmy Wu, Guanzhi Wang, S.~Shankar Sastry, Ken Goldberg, Linxi~"Jim" Fan, Yuke Zhu, and Guanya Shi.
\newblock {ENPIRE}: Agentic robot policy self-improvement in the real world, 2026.
\newblock URL \url{https://arxiv.org/abs/2606.19980}.

\bibitem[Xu et~al.(2023)Xu, Huang, Yu, Liu, Zhang, Niu, Zhang, Xia, Tan, and Zhao]{xu2023creative}
Mengdi Xu, Peide Huang, Wenhao Yu, Shiqi Liu, Xilun Zhang, Yaru Niu, Tingnan Zhang, Fei Xia, Jie Tan, and Ding Zhao.
\newblock Creative robot tool use with large language models.
\newblock \emph{arXiv preprint arXiv:2310.13065}, 2023.

\bibitem[Yin et~al.(2026)Yin, Ge, Wang, Wang, Li, Black, Darrell, Kanazawa, and Feng]{yin2026visionasinversegraphicsagentinterleavedmultimodal}
Shaofeng Yin, Jiaxin Ge, Zora~Zhiruo Wang, Chenyang Wang, Xiuyu Li, Michael~J. Black, Trevor Darrell, Angjoo Kanazawa, and Haiwen Feng.
\newblock Vision-as-inverse-graphics agent via interleaved multimodal reasoning, 2026.
\newblock URL \url{https://arxiv.org/abs/2601.11109}.

\bibitem[Yu et~al.(2023)Yu, Gileadi, Fu, Kirmani, Lee, Arenas, Chiang, Erez, Hasenclever, Humplik, Ichter, Xiao, Xu, Zeng, Zhang, Heess, Sadigh, Tan, Tassa, and Xia]{yu2023l2r}
Wenhao Yu, Nimrod Gileadi, Chuyuan Fu, Sean Kirmani, Kuang-Huei Lee, Montserrat~Gonzalez Arenas, Hao-Tien~Lewis Chiang, Tom Erez, Leonard Hasenclever, Jan Humplik, Brian Ichter, Ted Xiao, Peng Xu, Andy Zeng, Tingnan Zhang, Nicolas Heess, Dorsa Sadigh, Jie Tan, Yuval Tassa, and Fei Xia.
\newblock Language to rewards for robotic skill synthesis.
\newblock In Jie Tan, Marc Toussaint, and Kourosh Darvish (eds.), \emph{Proceedings of The 7th Conference on Robot Learning}, volume 229 of \emph{Proceedings of Machine Learning Research}, pp.\  374--404. PMLR, 06--09 Nov 2023.
\newblock URL \url{https://proceedings.mlr.press/v229/yu23a.html}.

\bibitem[Zhang et~al.(2023)Zhang, Zhang, Pertsch, Liu, Ren, Chang, Sun, and Lim]{zhang2023bootstrap}
Jesse Zhang, Jiahui Zhang, Karl Pertsch, Ziyi Liu, Xiang Ren, Minsuk Chang, Shao-Hua Sun, and Joseph~J Lim.
\newblock Bootstrap your own skills: Learning to solve new tasks with large language model guidance.
\newblock \emph{arXiv preprint arXiv:2310.10021}, 2023.

\bibitem[Zhang et~al.(2026{\natexlab{a}})Zhang, Ge, Yoo, Fu, Yang, Liu, Saravanan, Yin, Yu, Niu, et~al.]{zhang2026playful}
Junyi Zhang, Jiaxin Ge, Hanjun Yoo, Letian Fu, Zihan Yang, Yaowei Liu, Raj Saravanan, Shaofeng Yin, Justin Yu, Dantong Niu, et~al.
\newblock Playful agentic robot learning.
\newblock \emph{arXiv preprint arXiv:2606.19419}, 2026{\natexlab{a}}.

\bibitem[Zhang et~al.(2026{\natexlab{b}})Zhang, Wang, Ouyang, Huang, Li, Su, Jin, Chai, Liang, Dou, Chen, and Chen]{zhang2026unexpectedrobotpolicyearly}
Wenbo Zhang, Kaixuan Wang, Yutao Ouyang, Xiaoyu Huang, Liyang Li, Kailun Su, Weiyang Jin, Wenhao Chai, Haotian Liang, Zhiyang Dou, Yue Chen, and Tianxing Chen.
\newblock An unexpected robot policy: Early evaluations of gpt-6 astra on robodojo and beyond, 2026{\natexlab{b}}.
\newblock URL \url{https://arxiv.org/abs/2609.24170}.

\bibitem[Zhang et~al.(2026{\natexlab{c}})Zhang, Zhang, Gao, Li, Liu, Zhu, Qiu, Yan, Liu, Tang, et~al.]{zhang2026harness}
Yixian Zhang, Huanming Zhang, Feng Gao, Xiao Li, Zhihao Liu, Chunyang Zhu, Jiaxing Qiu, Yuchen Yan, Jiyuan Liu, Wenhao Tang, et~al.
\newblock Harness vla: Steering frozen vlas into reliable manipulation primitives via memory-guided agents.
\newblock \emph{arXiv preprint arXiv:2607.08448}, 2026{\natexlab{c}}.

\bibitem[Zhou et~al.(2025)Zhou, Su, Chi, Zhang, Wang, Huang, Sheng, and Wang]{zhou2025code}
Enshen Zhou, Qi~Su, Cheng Chi, Zhizheng Zhang, Zhongyuan Wang, Tiejun Huang, Lu~Sheng, and He~Wang.
\newblock Code-as-monitor: Constraint-aware visual programming for reactive and proactive robotic failure detection.
\newblock In \emph{2025 IEEE/CVF Conference on Computer Vision and Pattern Recognition (CVPR)}, pp.\  6919--6929. IEEE, 2025.

\bibitem[Zucker et~al.(2013)Zucker, Ratliff, Dragan, Pivtoraiko, Klingensmith, Dellin, Bagnell, and Srinivasa]{zucker2013chomp}
Matt Zucker, Nathan Ratliff, Anca~D Dragan, Mihail Pivtoraiko, Matthew Klingensmith, Christopher~M Dellin, J~Andrew Bagnell, and Siddhartha~S Srinivasa.
\newblock Chomp: Covariant hamiltonian optimization for motion planning.
\newblock \emph{The International journal of robotics research}, 32\penalty0 (9-10):\penalty0 1164--1193, 2013.

\end{thebibliography}
\bibliographystyle{iclr2027_conference}

\clearpage

\appendix

\section{Implementation details}
\label{app:implementation}

\subsection{Hyperparameters}
\label{app:hyperparameters}

Table~\ref{tab:hyperparameters} lists the default settings used in the main
experiments.

\begin{table}[h]
\caption{Default hyperparameters used in the main experiments. Population
size, per-process pretraining budget, deployment-trial budget, and evaluation
protocol are matched across methods whenever the corresponding phase applies.}
\label{tab:hyperparameters}
\centering
\small
\setlength{\tabcolsep}{5pt}
\begin{tabular}{@{}p{0.18\linewidth}p{0.58\linewidth}p{0.12\linewidth}@{}}
\toprule
\textbf{Phase} & \textbf{Hyperparameter} & \textbf{Value} \\
\midrule
Stage~1 \& 2 & Coding-agent orchestrator & Fable~5.1 \\
Stage~1 \& 2 & Policy-writing model & Fable~5.1 \\
Stage~1 & Reasoning effort & max \\
Stage~2 & Reasoning effort & medium \\
\midrule
Stage~1 & Independent population members & 4 \\
Stage~1 & Optimization iterations per member & 20 \\
Stage~1 & Broad probe configurations per iteration & 24--32 \\
Stage~1 & Focused configurations per selected sub-goal & $\sim$10 \\
Stage~1 & Capability-bank replay configurations per candidate & 4--6 \\
\midrule
Stage~2 & Robot rollouts per iteration & 1 \\
Stage~2 & Repair candidates per iteration & 4 \\
Stage~2 & Adaptation trials per run & 5 \\
Stage~2 & Agent-written test scenes per iteration & 5 \\
\midrule
Evaluation & Independent runs per method--task pair & 5 \\
Evaluation & I.i.d. evaluation trials per run & 5 \\
\bottomrule
\end{tabular}
\end{table}

\subsection{Policy-writing prompt}
\label{app:cap-prompt}

Every script executed in this paper, in simulation or on the robot, comes from
the same one-shot, code-as-policy call. The prompt below is shared by \method{},
every baseline, and every evaluation trial, across all three task families. It
concatenates the current \skillmd{}, the current \toolboxfile{} source, the
task, and a fixed output format. Slots shown in \pslot{\ldots} and
\pvar{italics} are filled in verbatim; everything else is literal. The call is
text-only: if the model uses a tool or takes more than one turn, the trial is
voided rather than retried.

\begin{promptbox}{Policy-writing (code-as-policy) prompt}
(*\pslot{skill.md}*)

(*\pmark{-{}-{}-{}-{}- TOOLBOX SOURCE (reference) -{}-{}-{}-{}-}*)
The current Toolbox implementation is below. Use it however you like:
call any helper, read what's available, or write your own code that
bypasses tb entirely. The only HARD rule is the no-cheating constraint
in the skill text above (no reading or writing of privileged object state).

(*\pslot{toolbox.py}*)

(*\pmark{-{}-{}-{}-{}- TASK -{}-{}-{}-{}-}*)
Name: (*\pvar{\$TASK\_NAME}*)
Description: (*\pvar{\$TASK\_DESC}*)

This is a pure text-generation call. Do not inspect files, do not run
shell commands, do not patch the filesystem, and do not use tools. The
skill text, toolbox source above, and task description contain all
required context.

(*\pmark{OUTPUT FORMAT}*) - overrides any conflicting instruction in skill.md:

(*\pmark{===PLAN\_START===}*)
concise plan, <=200 words
(*\pmark{===PLAN\_END===}*)

(*\pmark{===SCRIPT\_START===}*)
complete Python script, no markdown fences
(*\pmark{===SCRIPT\_END===}*)

Output ONLY the marker blocks. Do NOT call any tools.
\end{promptbox}

\subsection{Agent thinking time and token usage}
\label{app:compute-cost}


Table~\ref{tab:compute-cost} reports the cost of the orchestrating coding
agent on average for the barcode scanning tasks. 
Agent
thinking time sums, over all orchestrator calls, the time from the tool result
or prompt to the model's reply. Output tokens include reasoning tokens. Note that these costs are for toolbox optimization: once the toolbox is optimized, running inference for a given task takes only around 30 seconds.

\begin{table}[h]
\caption{Coding agent orchestration cost.}
\label{tab:compute-cost}
\centering
\small
\begin{tabular}{@{}lcc@{}}
\toprule
 & Stage~1 (per member) & Stage~2 (per run) \\
\midrule
Orchestrator calls & 840 & 124 \\
Agent thinking time & 5.4\,h & 31\,min \\
Output tokens (reasoning) & 1.98M (1.16M) & 161k (64k) \\
Cache-read input tokens & 155M & 19.5M \\
\bottomrule
\end{tabular}
\end{table}

\section{Evaluation task details}
\label{app:tasks}

\begin{figure}[H]
\centering
\includegraphics[width=\textwidth]{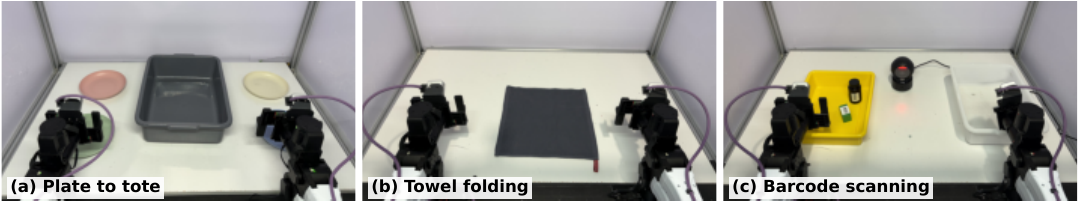}
\caption{Physical setups for the three evaluation task families: plate to
tote, towel folding, and barcode scanning. All use the same dual-arm YAM
platform and workspace enclosure, while task-specific objects and fixtures
produce distinct manipulation challenges.}
\label{fig:scenes}
\end{figure}

\subsection{Plate to tote}

\begin{figure}[H]
\centering
\includegraphics[width=\textwidth]{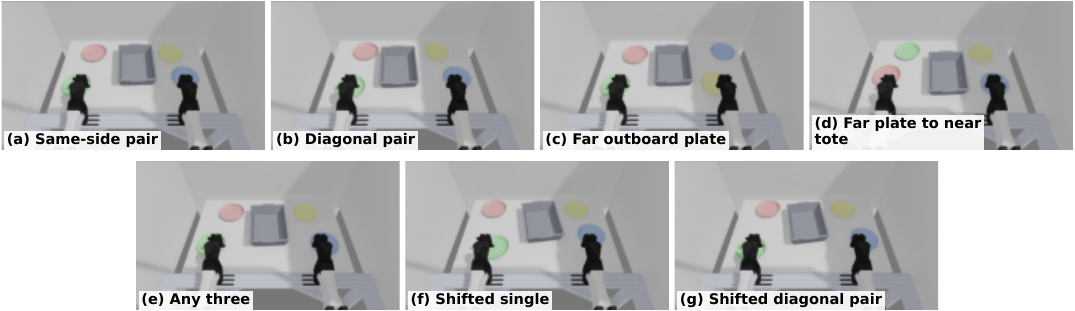}
\caption{Initial states of the seven plate to tote tasks. (Isaac Sim)}
\label{fig:tasks-plate}
\end{figure}

The robot must move one or more plates from a table into a tote. A plate lying flat cannot be picked up with a straightforward top-down grasp.
The task therefore tests whether the coding agent can discover a geometry-aware
strategy, such as a lateral rim pinch, and turn it into collision-free grasp,
lift, and transport motions.

Every delivered plate must first be lifted at least 5\,cm and end inside the
tote; plates not named by the task must remain outside.
\begin{itemize}[leftmargin=1.5em]
  \item \textbf{Same-side pair.} Move the near green and far red plates
  from the left side into the tote, one after the other.
  \item \textbf{Diagonal pair.} Move the near-left green and far-right
  yellow plates, requiring both arms and two distinct reaches.
  \item \textbf{Far outboard plate.} Move the blue plate from the far,
  outboard edge of the right arm's workspace.
  \item \textbf{Far plate to near tote.} Move a far-left green plate into
  a tote shifted toward the robot, producing a long return carry.
  \item \textbf{Any three.} Choose and move any three of the four plates;
  exactly one may remain on the table.
  \item \textbf{Shifted single.} Move the near-left green plate after the
  tote is translated 3.5\,cm, shifted laterally by 3\,cm, and rotated
  $6\degree$; all plate positions are also perturbed.
  \item \textbf{Shifted diagonal pair.} Repeat the green--yellow diagonal
  delivery under the same shifted fixture layout.
\end{itemize}

\subsection{Towel folding}

\begin{figure}[H]
\centering
\includegraphics[width=\textwidth]{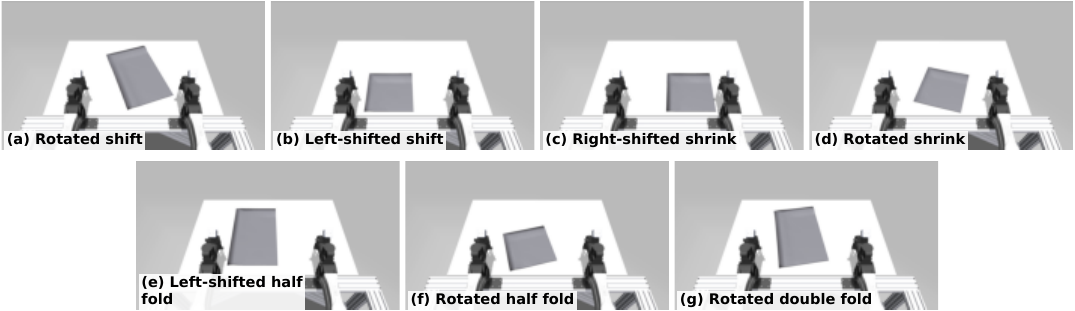}
\caption{Initial states of the seven towel folding tasks. (mjlab)}
\label{fig:tasks-towel}
\end{figure}

The robot must bring a deformable towel from varied initial poses into a target state, such as shifted to a new location, compacted into a smaller footprint, or folded once or twice.
Deformable-object dynamics make the outcome highly sensitive to grasp
location, drag trajectory, and accumulated error across folds, creating a
large gap between the sandbox and deployment environment.

\begin{itemize}[leftmargin=1.5em]
  \item \textbf{Rotated shift.} Move the center of a flat 56$\times$36\,cm
  towel, initially rotated by $20\degree$, by at least 5\,cm.
  \item \textbf{Left-shifted shift.} A 34\,cm square towel starts 10\,cm left
  of its nominal pose and its center must move by at least 5\,cm.
  \item \textbf{Right-shifted shrink.} The 34\,cm towel starts 10\,cm right of
  nominal; reduce its covered area to at most 80\% of the flat footprint.
  \item \textbf{Rotated shrink.} Perform the same shrink on the 34\,cm towel,
  rotated by $15\degree$ and placed 3\,cm farther from the robot.
  \item \textbf{Left-shifted half fold.} Fold a towel shifted 12\,cm left so
  its footprint is at most 62\% of the flat footprint along one axis.
  \item \textbf{Rotated half fold.} Fold the 34\,cm towel, rotated
  by $15\degree$ and placed 8\,cm left and 3\,cm farther, in half. Success
  requires a footprint of at most 62\% of flat and a half-fold shape---one side
  at most 62\% and the other at least 78\% of its flat length---with the towel
  lying flat; piled configurations fail.
  \item \textbf{Rotated double fold.} Starting 8\,cm left and rotated by
  $10\degree$, fold twice until the footprint is at most 38\% of flat.
\end{itemize}

\subsection{Barcode scanning}

\begin{figure}[H]
\centering
\includegraphics[width=\textwidth]{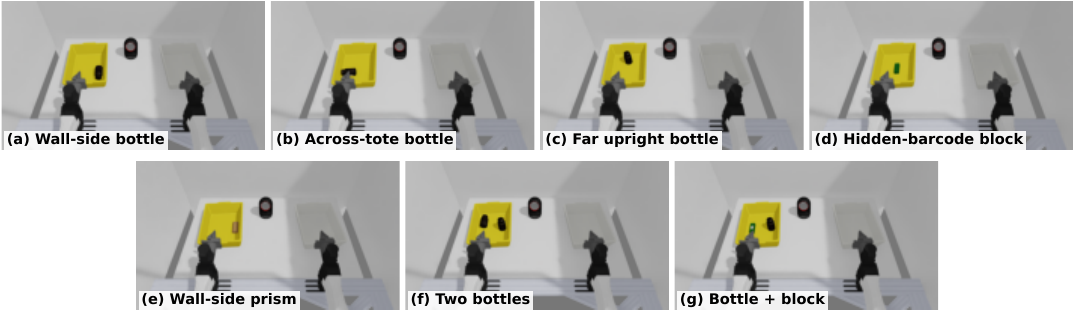}
\caption{Initial states of the seven barcode scanning tasks. (Isaac Sim)}
\label{fig:tasks-scan}
\end{figure}

The robot moves labeled objects from one tote to another after presenting each
label to a fixed barcode reader. This is a long-horizon task with a strict
logical sequence: the robot must discover the objects, acquire a feasible
grasp, expose the correct label face, align it with the reader, and only then
stow the object. The reader has a narrow operating window, tote walls
constrain the approach to objects near an edge, and multi-object variants
require reasoning about clutter and manipulation order.

\begin{itemize}[leftmargin=1.5em]
  \item \textbf{Wall-side bottle.} A bottle lies beside the
  near wall with its barcode facing into the wall. The policy must extract or
  roll it, present the barcode to the fixed scanner, and stow it.
  \item \textbf{Across-tote bottle.} The bottle spans the tote,
  with its cap close to the far wall and its barcode facing away. Success
  requires a clearance-aware grasp, reorientation, scanning, and stowing.
  \item \textbf{Far upright bottle.} An upright bottle starts near the
  far-reach boundary with its barcode already visible. The challenge is the
  long reach and subsequent transfer to the scanner and drop-off tote.
  \item \textbf{Hidden-barcode block.} A green block
  lies in open space with its printed face against the floor. It must be
  flipped or regrasped before it can be scanned and stowed.
  \item \textbf{Wall-side prism.} A flattened prism lies beside the near wall with its barcode against the floor.
  The policy must combine a wall-tight grasp with flipping, scanning, and
  stowing.
  \item \textbf{Two bottles.} One bottle is upright and one lies along a
  wall. Both barcodes must be read and both bottles must end in the drop-off
  tote; order is unconstrained.
  \item \textbf{Bottle + block.} The scene contains an upright bottle
  and a flat, barcode-up block. Both objects must be identified, scanned, and
  stowed, requiring shape-specific grasps and multi-object sequencing.
\end{itemize}

\section{Baseline details}
\label{app:baselines}

\paragraph{Controlled comparison.}
All methods start from the same low-level robot API
(Section~\ref{sec:library}) and initial toolbox (Section~\ref{sec:prep}), including the same raw camera observations,
calibration, kinematic queries, and generic motion wrappers; none receives a
predefined task-level manipulation skill. We hold the coding agent, reasoning
effort, task seeds, and evaluation protocol fixed across different methods. Every method with simulated pretraining
uses a four-member population with 20 optimization iterations per member,
while every method with real-world adaptation receives five deployment trials.  All
simulation-based methods also start from the same initial sandbox constructed
in Section~\ref{sec:prep} for each task family. The comparison therefore
isolates how each method utilizes simulation and robot experience rather than
differences in initialization or compute.

\paragraph{Setting-matched baselines.}
Our evaluation setting combines code-policy generation, a persistent toolbox,
open-ended simulation pretraining, and few-trial deployment adaptation. Since
some prior methods were introduced under slightly different setups, we mark a
baseline with an asterisk when we preserve its central mechanism but
reimplement it under our common interface and budget. The asterisk denotes a
setting-matched implementation, not additional information or a relaxed
evaluation.

\paragraph{Methods.}
\begin{itemize}[leftmargin=1.5em]
  \item \textbf{Direct CaP} independently generates a task program
  from the shared initial toolbox and low-level robot API for each trial,
  following Code as Policies \citep{liang2023code}.
  \item \textbf{ENPIRE$^*$} starts from the same initial toolbox and iteratively
  edits the policy using only deployment rollouts. It instantiates ENPIRE's
  reset--execute--verify--refine loop \citep{xiao2026enpire}; the asterisk
  denotes our code-policy implementation under the shared five-trial budget.
  \item \textbf{Zero-Shot Sim2Real} uses a coding agent to optimize the
  toolbox entirely through simulated experiments using our Stage 1 pipeline, then transfers it directly
  to the target environment without deployment-time adaptation. This baseline
  follows the simulation-only transfer principle of DrEureka
  \citep{ma2024dreureka}, instantiated through the shared code-policy interface.
  \item \textbf{ASPIRE$^*$} builds a textual skill library in simulation and
  uses it for deployment-time repair, following ASPIRE
  \citep{lu2026aspire}. The asterisk marks our interface- and budget-matched
  implementation.
  \item \textbf{\method{} (no discovery)} replaces
  Stage~1 with a fixed-target autoresearch run
  \citep{karpathy2026autoresearch}, then applies the same Stage~2 as \method{}.
  \item \textbf{\method{} (direct repair)} starts from the same
  Stage~1 toolbox as \method{}, but performs Stage~2 repair directly from each deployment
  rollout, without simulator replay, hypothesis generation, or
  candidate screening.
\end{itemize}

\section{Additional results}
\label{app:results}

\subsection{Per-task sim-to-sim results}
\label{app:pertask}

Table~\ref{tab:pertask-results} reports the per-task counts behind
Figure~\ref{fig:scene-baselines}. Each table cell reports successful
evaluation trials over 25 scheduled trials; bold marks the highest success
proportion in each row.

\begin{table}[h]
\caption{Per-task controlled sim-to-sim results, reported as successful
evaluation trials out of 25.}
\label{tab:pertask-results}
\centering
\setlength{\tabcolsep}{3pt}
\resizebox{\textwidth}{!}{%
\begin{tabular}{@{}llccccccc@{}}
\toprule
Scene & Task & \shortstack{Direct\\CaP} & ENPIRE$^*$ &
\shortstack{Zero-Shot\\Sim2Real} & ASPIRE$^*$ &
\shortstack{\method{}\\(no discovery)} &
\shortstack{\method{}\\(direct repair)} &
\textbf{\shortstack{\method{}\\(ours)}} \\
\midrule
\multirow{7}{*}{\shortstack{Plate\\to tote}}
& Same-side pair        & \pcnt{1}{25} & \pcnt{4}{25} & \pcnt{1}{25} & \pcnt{0}{25} & \pcnt{16}{25} & \bpcnt{21}{25} & \bpcnt{21}{25} \\
& Diagonal pair         & \pcnt{3}{25} & \pcnt{8}{25} & \pcnt{0}{25} & \pcnt{0}{25} & \pcnt{12}{25} & \pcnt{14}{25} & \bpcnt{24}{25} \\
& Far outboard plate    & \pcnt{17}{25} & \pcnt{14}{25} & \pcnt{12}{25} & \pcnt{17}{25} & \pcnt{15}{25} & \pcnt{19}{25} & \bpcnt{22}{25} \\
& Far plate to near tote & \pcnt{9}{25} & \pcnt{18}{25} & \pcnt{10}{25} & \pcnt{19}{25} & \pcnt{20}{25} & \bpcnt{25}{25} & \bpcnt{25}{25} \\
& Any three             & \pcnt{0}{25} & \pcnt{6}{25} & \pcnt{0}{25} & \pcnt{0}{25} & \bpcnt{23}{25} & \pcnt{16}{25} & \bpcnt{23}{25} \\
& Shifted single        & \pcnt{11}{25} & \pcnt{5}{25} & \bpcnt{25}{25} & \pcnt{16}{25} & \pcnt{6}{25} & \pcnt{23}{25} & \bpcnt{25}{25} \\
& Shifted diagonal pair & \pcnt{3}{25} & \pcnt{7}{25} & \pcnt{0}{25} & \pcnt{0}{25} & \pcnt{13}{25} & \pcnt{20}{25} & \bpcnt{25}{25} \\
\midrule
\multirow{7}{*}{\shortstack{Towel\\folding}}
& Rotated shift          & \pcnt{0}{25} & \pcnt{8}{25} & \pcnt{5}{25} & \pcnt{0}{25} & \pcnt{16}{25} & \pcnt{5}{25} & \bpcnt{21}{25} \\
& Left-shifted shift     & \pcnt{4}{25} & \pcnt{4}{25} & \pcnt{5}{25} & \pcnt{11}{25} & \pcnt{14}{25} & \pcnt{10}{25} & \bpcnt{19}{25} \\
& Right-shifted shrink   & \pcnt{5}{25} & \pcnt{14}{25} & \pcnt{17}{25} & \pcnt{20}{25} & \pcnt{19}{25} & \pcnt{15}{25} & \bpcnt{24}{25} \\
& Rotated shrink         & \pcnt{10}{25} & \pcnt{16}{25} & \pcnt{12}{25} & \pcnt{18}{25} & \pcnt{20}{25} & \pcnt{19}{25} & \bpcnt{21}{25} \\
& Left-shifted half fold & \pcnt{7}{25} & \pcnt{7}{25} & \pcnt{8}{25} & \bpcnt{11}{25} & \pcnt{9}{25} & \pcnt{10}{25} & \bpcnt{11}{25} \\
& Rotated half fold      & \pcnt{1}{25} & \pcnt{3}{25} & \pcnt{5}{25} & \pcnt{3}{25} & \pcnt{2}{25} & \pcnt{5}{25} & \bpcnt{8}{25} \\
& Rotated double fold    & \pcnt{3}{25} & \pcnt{8}{25} & \pcnt{11}{25} & \pcnt{4}{25} & \pcnt{2}{25} & \pcnt{4}{25} & \bpcnt{14}{25} \\
\midrule
\multirow{7}{*}{\shortstack{Barcode\\scanning}}
& Wall-side bottle & \pcnt{0}{25} & \pcnt{0}{25} & \pcnt{0}{25} & \pcnt{0}{25} & \pcnt{2}{25} & \pcnt{17}{25} & \bpcnt{25}{25} \\
& Across-tote bottle & \pcnt{0}{25} & \pcnt{0}{25} & \pcnt{0}{25} & \pcnt{1}{25} & \pcnt{7}{25} & \pcnt{17}{25} & \bpcnt{24}{25} \\
& Far upright bottle & \pcnt{0}{25} & \pcnt{0}{25} & \pcnt{0}{25} & \pcnt{15}{25} & \pcnt{17}{25} & \pcnt{16}{25} & \bpcnt{24}{25} \\
& Hidden-barcode block & \pcnt{1}{25} & \pcnt{0}{25} & \pcnt{0}{25} & \pcnt{10}{25} & \pcnt{8}{25} & \bpcnt{25}{25} & \pcnt{24}{25} \\
& Wall-side prism & \pcnt{0}{25} & \pcnt{0}{25} & \pcnt{0}{25} & \pcnt{6}{25} & \pcnt{6}{25} & \pcnt{3}{25} & \bpcnt{19}{25} \\
& Two bottles        & \pcnt{0}{25} & \pcnt{0}{25} & \pcnt{2}{25} & \pcnt{0}{25} & \pcnt{2}{25} & \bpcnt{20}{25} & \pcnt{19}{25} \\
& Bottle + block     & \pcnt{0}{25} & \pcnt{0}{25} & \pcnt{0}{25} & \pcnt{0}{25} & \pcnt{0}{25} & \pcnt{4}{25} & \bpcnt{14}{25} \\
\bottomrule
\end{tabular}}
\end{table}

\subsection{Per-task ablation results}
\label{app:ablations}

For Stage~1, \emph{No population} uses one optimization process instead of
four, while \emph{No capability bank} removes capability-bank replay. For
Stage~2, \emph{No repair alternatives} limits each iteration to one proposed
repair, removing comparison among alternatives, while \emph{No hypothesis
generation} formulates repairs directly from the robot evidence without first
stating hypotheses. Each setting uses five runs per task and five evaluation
trials per run. Table~\ref{tab:ablations} reports the per-task counts behind
Figure~\ref{fig:ablations}. 

\begin{table}[h]
\caption{Per-task barcode scanning ablations, reported as successful evaluation
trials out of 25.}
\label{tab:ablations}
\centering
\resizebox{\linewidth}{!}{%
\begin{tabular}{@{}lccccc@{}}
\toprule
Task & \method{} (ours) & \shortstack{No population\\(Stage~1)} &
\shortstack{No capability bank\\(Stage~1)} &
\shortstack{No repair alternatives\\(Stage~2)} &
\shortstack{No hypothesis generation\\(Stage~2)} \\
\midrule
Wall-side bottle & \pcnt{25}{25} & \pcnt{17}{25} & \pcnt{15}{25} & \pcnt{24}{25} & \pcnt{23}{25} \\
Across-tote bottle & \pcnt{24}{25} & \pcnt{20}{25} & \pcnt{14}{25} & \pcnt{25}{25} & \pcnt{20}{25} \\
Far upright bottle & \pcnt{24}{25} & \pcnt{25}{25} & \pcnt{25}{25} & \pcnt{25}{25} & \pcnt{20}{25} \\
Hidden-barcode block & \pcnt{24}{25} & \pcnt{25}{25} & \pcnt{24}{25} & \pcnt{25}{25} & \pcnt{25}{25} \\
Wall-side prism & \pcnt{19}{25} & \pcnt{15}{25} & \pcnt{21}{25} & \pcnt{5}{25} & \pcnt{13}{25} \\
Two bottles        & \pcnt{19}{25} & \pcnt{12}{25} & \pcnt{15}{25} & \pcnt{18}{25} & \pcnt{16}{25} \\
Bottle + block     & \pcnt{14}{25} & \pcnt{1}{25} & \pcnt{7}{25} & \pcnt{14}{25} & \pcnt{9}{25} \\
\bottomrule
\end{tabular}}
\end{table}

\subsection{Per-task results across coding agents}
\label{app:cross-agent-results}

\begin{figure}[h]
\centering
\includegraphics[width=0.6\linewidth]{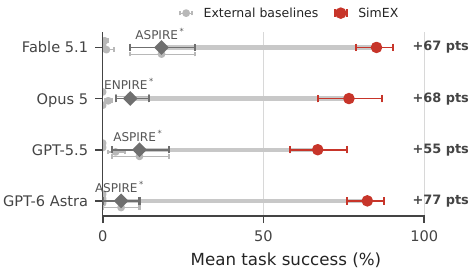}
\captionof{figure}{Across all coding agents, \method{} consistently outperforms the strongest baseline methods (marked with darker diamond) by a large margin.}
\label{fig:cross-agent-gap}
\end{figure}

Tables~\ref{tab:cross-agent-f51}--\ref{tab:cross-agent-g6a} report the
complete per-task results underlying
Figures~\ref{fig:cross-agent-heatmap}--\ref{fig:cross-agent-gap}. Each
configuration uses the same seven tasks, five independent runs, and five
evaluation trials per run. Bold marks the highest success proportion in each
row.

\begin{table*}[!t]
\caption{Per-task barcode scanning results for the all-Fable~5.1 configuration used in the main experiments, reported as successful evaluation trials out of 25.}
\label{tab:cross-agent-f51}
\centering
\small
\setlength{\tabcolsep}{4.5pt}
\resizebox{\textwidth}{!}{%
\begin{tabular}{@{}lccccccc@{}}
\toprule
Task & \shortstack{Direct\\CaP} & ENPIRE$^*$ & \shortstack{Zero-Shot\\Sim2Real} & ASPIRE$^*$ & \shortstack{\method{}\\(no discovery)} & \shortstack{\method{}\\(direct repair)} & \textbf{\method{} (ours)} \\
\midrule
Wall-side bottle & \pcnt{0}{25} & \pcnt{0}{25} & \pcnt{0}{25} & \pcnt{0}{25} & \pcnt{2}{25} & \pcnt{17}{25} & \bpcnt{25}{25} \\
Across-tote bottle & \pcnt{0}{25} & \pcnt{0}{25} & \pcnt{0}{25} & \pcnt{1}{25} & \pcnt{7}{25} & \pcnt{17}{25} & \bpcnt{24}{25} \\
Far upright bottle & \pcnt{0}{25} & \pcnt{0}{25} & \pcnt{0}{25} & \pcnt{15}{25} & \pcnt{17}{25} & \pcnt{16}{25} & \bpcnt{24}{25} \\
Hidden-barcode block & \pcnt{1}{25} & \pcnt{0}{25} & \pcnt{0}{25} & \pcnt{10}{25} & \pcnt{8}{25} & \bpcnt{25}{25} & \pcnt{24}{25} \\
Wall-side prism & \pcnt{0}{25} & \pcnt{0}{25} & \pcnt{0}{25} & \pcnt{6}{25} & \pcnt{6}{25} & \pcnt{3}{25} & \bpcnt{19}{25} \\
Two bottles & \pcnt{0}{25} & \pcnt{0}{25} & \pcnt{2}{25} & \pcnt{0}{25} & \pcnt{2}{25} & \bpcnt{20}{25} & \pcnt{19}{25} \\
Bottle + block & \pcnt{0}{25} & \pcnt{0}{25} & \pcnt{0}{25} & \pcnt{0}{25} & \pcnt{0}{25} & \pcnt{4}{25} & \bpcnt{14}{25} \\
\bottomrule
\end{tabular}}
\end{table*}

\begin{table*}[!t]
\caption{Per-task barcode scanning results for the end-to-end Opus~5 configuration, reported as successful evaluation trials out of 25.}
\label{tab:cross-agent-o5}
\centering
\small
\setlength{\tabcolsep}{4.5pt}
\resizebox{\textwidth}{!}{%
\begin{tabular}{@{}lccccccc@{}}
\toprule
Task & \shortstack{Direct\\CaP} & ENPIRE$^*$ & \shortstack{Zero-Shot\\Sim2Real} & ASPIRE$^*$ & \shortstack{\method{}\\(no discovery)} & \shortstack{\method{}\\(direct repair)} & \textbf{\method{} (ours)} \\
\midrule
Wall-side bottle & \pcnt{0}{25} & \pcnt{4}{25} & \pcnt{0}{25} & \pcnt{0}{25} & \pcnt{0}{25} & \pcnt{16}{25} & \bpcnt{25}{25} \\
Across-tote bottle & \pcnt{0}{25} & \pcnt{1}{25} & \pcnt{0}{25} & \pcnt{0}{25} & \pcnt{3}{25} & \pcnt{20}{25} & \bpcnt{25}{25} \\
Far upright bottle & \pcnt{0}{25} & \pcnt{8}{25} & \pcnt{0}{25} & \pcnt{0}{25} & \pcnt{7}{25} & \pcnt{20}{25} & \bpcnt{22}{25} \\
Hidden-barcode block & \pcnt{0}{25} & \pcnt{2}{25} & \pcnt{0}{25} & \pcnt{0}{25} & \pcnt{17}{25} & \bpcnt{24}{25} & \pcnt{23}{25} \\
Wall-side prism & \pcnt{0}{25} & \pcnt{0}{25} & \pcnt{0}{25} & \pcnt{0}{25} & \pcnt{4}{25} & \pcnt{2}{25} & \bpcnt{10}{25} \\
Two bottles & \pcnt{0}{25} & \pcnt{0}{25} & \pcnt{3}{25} & \pcnt{0}{25} & \pcnt{1}{25} & \pcnt{21}{25} & \bpcnt{23}{25} \\
Bottle + block & \pcnt{0}{25} & \pcnt{0}{25} & \pcnt{0}{25} & \pcnt{0}{25} & \pcnt{0}{25} & \pcnt{5}{25} & \bpcnt{6}{25} \\
\bottomrule
\end{tabular}}
\end{table*}

\begin{table*}[!t]
\caption{Per-task barcode scanning results for the end-to-end GPT-5.5 configuration, reported as successful evaluation trials out of 25.}
\label{tab:cross-agent-g55}
\centering
\small
\setlength{\tabcolsep}{4.5pt}
\resizebox{\textwidth}{!}{%
\begin{tabular}{@{}lccccccc@{}}
\toprule
Task & \shortstack{Direct\\CaP} & ENPIRE$^*$ & \shortstack{Zero-Shot\\Sim2Real} & ASPIRE$^*$ & \shortstack{\method{}\\(no discovery)} & \shortstack{\method{}\\(direct repair)} & \textbf{\method{} (ours)} \\
\midrule
Wall-side bottle & \pcnt{0}{25} & \pcnt{0}{25} & \pcnt{0}{25} & \pcnt{0}{25} & \pcnt{2}{25} & \pcnt{19}{25} & \bpcnt{21}{25} \\
Across-tote bottle & \pcnt{0}{25} & \pcnt{0}{25} & \pcnt{0}{25} & \pcnt{9}{25} & \pcnt{1}{25} & \pcnt{13}{25} & \bpcnt{21}{25} \\
Far upright bottle & \pcnt{0}{25} & \pcnt{0}{25} & \pcnt{0}{25} & \pcnt{1}{25} & \pcnt{0}{25} & \pcnt{10}{25} & \bpcnt{17}{25} \\
Hidden-barcode block & \pcnt{0}{25} & \pcnt{0}{25} & \pcnt{0}{25} & \pcnt{5}{25} & \pcnt{15}{25} & \bpcnt{25}{25} & \pcnt{24}{25} \\
Wall-side prism & \pcnt{0}{25} & \pcnt{0}{25} & \pcnt{0}{25} & \bpcnt{5}{25} & \pcnt{4}{25} & \pcnt{0}{25} & \bpcnt{5}{25} \\
Two bottles & \pcnt{0}{25} & \pcnt{0}{25} & \pcnt{7}{25} & \pcnt{0}{25} & \pcnt{0}{25} & \bpcnt{25}{25} & \pcnt{24}{25} \\
Bottle + block & \pcnt{0}{25} & \pcnt{0}{25} & \pcnt{0}{25} & \pcnt{0}{25} & \pcnt{0}{25} & \bpcnt{5}{25} & \bpcnt{5}{25} \\
\bottomrule
\end{tabular}}
\end{table*}

\begin{table*}[!t]
\caption{Per-task barcode scanning results for the end-to-end GPT-6 Astra configuration, reported as successful evaluation trials out of 25.}
\label{tab:cross-agent-g6a}
\centering
\small
\setlength{\tabcolsep}{4.5pt}
\resizebox{\textwidth}{!}{%
\begin{tabular}{@{}lccccccc@{}}
\toprule
Task & \shortstack{Direct\\CaP} & ENPIRE$^*$ & \shortstack{Zero-Shot\\Sim2Real} & ASPIRE$^*$ & \shortstack{\method{}\\(no discovery)} & \shortstack{\method{}\\(direct repair)} & \textbf{\method{} (ours)} \\
\midrule
Wall-side bottle & \pcnt{0}{25} & \pcnt{0}{25} & \pcnt{0}{25} & \pcnt{0}{25} & \pcnt{4}{25} & \pcnt{17}{25} & \bpcnt{25}{25} \\
Across-tote bottle & \pcnt{0}{25} & \pcnt{0}{25} & \pcnt{0}{25} & \pcnt{0}{25} & \pcnt{0}{25} & \pcnt{17}{25} & \bpcnt{24}{25} \\
Far upright bottle & \pcnt{0}{25} & \pcnt{0}{25} & \pcnt{0}{25} & \pcnt{0}{25} & \pcnt{6}{25} & \pcnt{20}{25} & \bpcnt{25}{25} \\
Hidden-barcode block & \pcnt{0}{25} & \pcnt{0}{25} & \pcnt{0}{25} & \pcnt{10}{25} & \pcnt{9}{25} & \bpcnt{25}{25} & \bpcnt{25}{25} \\
Wall-side prism & \pcnt{0}{25} & \pcnt{0}{25} & \pcnt{0}{25} & \pcnt{0}{25} & \pcnt{0}{25} & \pcnt{0}{25} & \bpcnt{8}{25} \\
Two bottles & \pcnt{0}{25} & \pcnt{0}{25} & \pcnt{0}{25} & \pcnt{0}{25} & \pcnt{0}{25} & \pcnt{17}{25} & \bpcnt{20}{25} \\
Bottle + block & \pcnt{0}{25} & \pcnt{0}{25} & \pcnt{0}{25} & \pcnt{0}{25} & \pcnt{0}{25} & \pcnt{0}{25} & \bpcnt{17}{25} \\
\bottomrule
\end{tabular}}
\end{table*}

\subsection{Comparison with online agent control}
\label{app:online-control}

An alternative to synthesizing a robot program is to use the coding agent
itself as an online policy~\citep{zhang2026unexpectedrobotpolicyearly}: after each observation, the agent emits the next
low-level action. This approach avoids committing to a program before
execution, but places serial model inference directly on the robot's control
path. We compare these two deployment modes on three representative
barcode scanning tasks. For \method{}, the coding agent writes one task program
before execution. For the online policy, the same coding-agent model observes
camera-derived object locations and proprioception, then repeatedly emits
end-effector actions. We evaluate the online policy over five initializations
per task.

\method{} makes a single model call per task, before the task begins. This
call takes about 30 seconds, and all three generated programs complete their
validation rollout. Execution itself contains no model calls.
Moreover, when the task interface is unchanged, the same program can be reused
across initial states, allowing the synthesis cost to be amortized.

In contrast, the online policy makes about 30 model calls per rollout and
waits 40 minutes for their responses, over 50 times longer than it spends
executing actions. None of its 15 rollouts completes the task within the
allotted wall-clock budget. The bottleneck is therefore not robot
motion, but repeated serial inference while the robot waits for its next
command.
Although future models may improve both latency and control quality, moving
coding-agent inference out of the execution loop makes the \method{} system
substantially more practical for robotics.

\section{Extended related work}
\label{app:related}

In this section, we review four areas most closely related to our method:
LLM-based robot control, agentic optimization, generative simulation and
environment design, and sim-to-real transfer.

\paragraph{LLM-based Robot Control.}
Prior work uses LLMs to map language instructions to robot behavior by
generating executable plans or code, incorporating language feedback, defining
optimization objectives, or orchestrating learned policies
\citep{liang2023code,singh2023progprompt,huang2022inner,yu2023l2r,hu2026mattersorchestratingrobotpolicies,chen2026show,zhang2026harness,fu2026cap,jia2026agent,vemprala2024chatgpt,liu2024moka,zhou2025code,mu2024robocodex,xu2023creative}.
These methods demonstrate that an LLM can compose robot behavior from a
provided interface, but generally assume that the underlying capabilities,
feedback channels, or optimization interface have already been specified.
\method{} instead studies how a coding agent can construct and calibrate these
underlying robot capabilities through experimentation.

\paragraph{Agentic Optimization.}
Autoresearch casts improvement as an iterative loop in which a coding agent
proposes, implements, and evaluates changes to an executable artifact
\citep{karpathy2026autoresearch}. In digital domains, this paradigm has been
used to discover algorithms and automate computational research
\citep{romera2024funsearch,novikov2025alphaevolve,lu2024aiscientist}. It has
also been extended to embodied settings through persistent skill libraries,
automatically optimized reward programs, and robot policies revised from
simulated or physical feedback
\citep{wang2023voyager,ma2024eureka,lu2026aspire,xiao2026enpire,zhang2026playful,zhang2023bootstrap}. \method{}
develops a robot capability library through open-ended simulated
autoresearch and then calibrates it with feedback from deployment. Its simulated
stage assumes neither a fixed task distribution, a prescribed set of
environment variations, nor a ground-truth evaluation metric.

\paragraph{Generative Simulation and Environment Design.}
Unsupervised environment design automatically constructs curricula or selects
environments that adapt to a learner~\citep{dennis2020emergent,jiang2021prioritized,parker2022evolving}, while recent generative simulation
frameworks use foundation models to produce simulator scenes, task
specifications, rewards, and training supervision~\citep{wang2024gensim,wang2023robogen,ha2023scaling,dai2024automated,torne2024reconciling}. Both directions reduce the
manual effort required to create diverse training environments and curricula.
They mainly use generated content to train policies. \method{} instead
generates task variations and success predicates in a simulation of the target
workspace to stress-test, diagnose, and expand a persistent symbolic code
toolbox.

\paragraph{Sim-to-Real Transfer.}
Because a simulator cannot exactly reproduce real-world perception, geometry,
and contact dynamics, policies developed in simulation may fail after physical
deployment. Sim-to-real transfer studies how to bridge this gap. Prior methods
improve transfer by exposing policies to varied simulation parameters,
or automating reward and randomization design
\citep{tobin2017domain,ma2024dreureka}. A particularly relevant design pattern
is simulation pretraining followed by limited real-world adaptation
\citep{hu2025slac,kumar2021rma,chebotar2019closing}. \method{} is
inspired by this two-stage structure, but differs in both the learned object
and the role of simulation: it constructs a reusable code toolbox through
open-ended autoresearch, then returns to simulation between physical trials to
diagnose failures and screen code repairs.

\end{document}